%% file: main.tex
\documentclass[10pt,twocolumn,letterpaper]{article}

\usepackage[pagenumbers]{cvpr}
\input{preamble}

\usepackage{colortbl}
\usepackage{needspace}
\usepackage{algorithm}
\usepackage{algpseudocode}
\usepackage{placeins}
\usepackage{fvextra}
\definecolor{cvprblue}{rgb}{0.21,0.49,0.74}
\usepackage[breaklinks,colorlinks,allcolors=cvprblue]{hyperref}

\title{PACT: Plan-Action Consistency Training by Preserving Correct Plans and Repairing Actions in GUI Agents}
\input{authors}
\hypersetup{
  pdftitle={PACT: Plan-Action Consistency Training by Preserving Correct Plans and Repairing Actions in GUI Agents},
  pdfauthor={\PACTAuthorMetadata},
  pdfsubject={Plan-Action Consistency Training for GUI Agents},
  pdfkeywords={GUI agents, reinforcement learning, credit assignment, plan-action consistency, action repair}
}

\begin{document}
\maketitle
\input{sec/0_abstract}
\input{sec/1_intro}
\input{sec/2_related_work}
\input{sec/3_problem}
\input{sec/4_method}
\input{sec/5_experiments}
\input{sec/6_results}
\input{sec/7_limitations}
\input{sec/8_conclusion}

\FloatBarrier

\paragraph{LLM Usage Statement.}
Large language models (GPT and Gemini) were used for manuscript copy-editing, figure-layout formatting, and assisting with initial candidate screening prior to human verification in our diagnostic semantic evaluations (Section~\ref{experimental-setup}). 
All experimental designs, implementations, analyses, and final judgments were conducted and verified by the authors.

{
    \small
    \bibliographystyle{ieeenat_fullname}
    \bibliography{main}
}

\clearpage
\raggedbottom
\setlength{\textfloatsep}{20pt plus 2pt minus 4pt}
\setlength{\floatsep}{12pt plus 2pt minus 2pt}
\setlength{\dbltextfloatsep}{20pt plus 2pt minus 4pt}
\setlength{\dblfloatsep}{12pt plus 2pt minus 2pt}
\maketitlesupplementary
\appendix
\setcounter{table}{0}
\setcounter{figure}{0}
\setcounter{equation}{0}
\setcounter{algorithm}{0}
\renewcommand{\thetable}{S\arabic{table}}
\renewcommand{\thefigure}{S\arabic{figure}}
\renewcommand{\theequation}{S\arabic{equation}}
\renewcommand{\thealgorithm}{S\arabic{algorithm}}
\renewcommand{\theHtable}{supp.\arabic{table}}
\renewcommand{\theHfigure}{supp.\arabic{figure}}
\renewcommand{\theHequation}{supp.\arabic{equation}}
\providecommand{\theHalgorithm}{\arabic{algorithm}}
\renewcommand{\theHalgorithm}{supp.\arabic{algorithm}}
\input{sec/X_suppl}
\end{document}

%% file: preamble.tex
\usepackage[T1]{fontenc}
\usepackage[utf8]{inputenc}
\usepackage{tabularx,array,multirow}
\usepackage{microtype}

%% file: authors.tex
\author{
Zhuohan Cai$^{1,2,*}$ \quad Zhangxuan Gu$^{2,\dagger}$ \quad Runze Li$^{2}$ \quad Beitong Zhou$^{2}$\\
Ming Wen$^{2}$ \quad Hongbo Yu$^{1,\dagger}$ \quad Shuheng Shen$^{2}$ \quad Changhua Meng$^{2}$\\[0.4em]
{\normalsize $^{1}$Tsinghua University \qquad $^{2}$Ant Group}\\[0.25em]
{\small\ttfamily caizh24@mails.tsinghua.edu.cn\quad guzhangxuan.gzx@antgroup.com\quad yuhongbo@mail.tsinghua.edu.cn}\\[0.25em]
{\small $^{*}$This work was done during an internship at Ant Group.}\\[0.15em]
{\small $^{\dagger}$Corresponding authors: Zhangxuan Gu and Hongbo Yu.}
}
\newcommand{\PACTAuthorMetadata}{Zhuohan Cai, Zhangxuan Gu, Runze Li, Beitong Zhou, Ming Wen, Hongbo Yu, Shuheng Shen, Changhua Meng}

%% file: sec/0_abstract.tex
\begin{abstract}
GUI agents can formulate a correct immediate plan in their reasoning
yet emit an action that fails to execute it, creating an
\emph{execution gap}.
This recurring mismatch undermines reliable task completion, as a
single execution error can derail a multi-step workflow.
In outcome-supervised reinforcement learning, broadcasting a
negative response-level advantage across all generated tokens risks
penalizing valid planning together with erroneous execution.
To address this credit-assignment conflict, we introduce
\textbf{PACT} (\textbf{P}lan--\textbf{A}ction \textbf{C}onsistency
\textbf{T}raining), a training framework that combines
reference-aligned plan protection and plan-aligned action repair
without altering the original scalar action reward.
Using annotated reference actions to identify execution gaps, PACT
protects the verified plan by masking negative policy-gradient
contributions across its token span. Keeping the generated reasoning
unchanged as context, it then favors the reference action over the
emitted erroneous action through a preference surrogate with
stop-gradient pair weights.
Compared with the outcome-only baseline, PACT raises
AndroidWorld task success from 60.34\% to \textbf{68.97\%}
(\textbf{+8.63\,pp}), while improving
rule-based plan--action consistency on AndroidControl High from
83.71\% to \textbf{89.74\%} (\textbf{+6.03\,pp}).
Adapting PACT to desktop agents also improves task success from
40.92\% to \textbf{44.17\%} (\textbf{+3.25\,pp}) on the full 369-task
OSWorld benchmark.
Together, these results establish PACT as an effective credit-assignment
framework for bridging execution gaps and improving task completion
across mobile and desktop GUI environments.
\end{abstract}

%% file: sec/1_intro.tex
\section{Introduction}\label{introduction}

GUI agents must translate visual interfaces and user goals into executable interactions. Yet formulating a correct immediate plan in generated reasoning does not guarantee that the emitted action executes it. For example, an agent may state that it will tap a visible \texttt{+} button, yet emit an application-launch command instead (Figure~\ref{fig:execution-gap}). This \emph{execution gap} is distinct from a planning error: the response already contains a valid immediate plan, but the final action fails to realize it. Building on prior diagnostic studies of reasoning--execution gaps in mobile agents \citep{dong2025gap}, we investigate how training can preserve valid plans within failed responses while repairing their execution.

\input{sec/fig1_execution_gap}

In reinforcement learning (RL), such execution gaps create a credit-assignment problem. Several GUI-agent methods use Group Relative Policy Optimization (GRPO) or its variants to learn from action feedback~\citep{gu2025uivenus,hu2026faithful}. In implementations that broadcast one response-level advantage across reasoning and action tokens~\citep{shao2024deepseekmath}, a negative advantage penalizes valid planning alongside erroneous execution. A scalar thought--action consistency reward~\citep{hu2026faithful} does not determine the direction of correction: higher agreement can result from adapting a plan to justify an incorrect action, and a flawed plan can be executed consistently. Effective correction therefore requires identifying which component the annotated reference action supports preserving.

Our central principle is to \textbf{preserve a reference-supported plan and repair the action that fails to implement it}. Here, a plan denotes the intended next GUI interaction expressed in the generated reasoning, and the plan span denotes the tokens expressing it. Verifying reference support for parameterized GUI actions requires checking spatial or textual arguments beyond action type alone, enabling selective credit assignment that shields valid plans while repairing failed actions.

We present \textbf{PACT}, which instantiates this principle within a group-relative RL pipeline while leaving the original scalar action reward intact. First, a conservative rule-based selector identifies responses featuring a reference-supported plan but an incorrect, inconsistent final action. Second, \emph{reference-aligned plan protection} masks negative policy-gradient contributions within the plan span, leaving the advantages on action tokens and the remaining reasoning unchanged. Third, \emph{plan-aligned action repair} pairs a reference-derived positive action suffix with the model's emitted erroneous suffix under the same decoded reasoning trace via a stop-gradient pair-weight preference surrogate in the spirit of direct preference optimization (DPO) \citep{rafailov2023dpo}, outside the original GRPO rollout group. Both mechanisms operate strictly during training; inference \mbox{retains} the standard single-pass reasoning--action interface without extra reflection or re-planning.

PACT improves interactive task completion over the original GRPO baseline, reaching 68.97\% success on AndroidWorld~\citep{rawles2024androidworld} (+8.63\,pp) and 44.17\% on the full OSWorld benchmark~\citep{xie2024osworld} (+3.25\,pp). It also achieves 89.74\% rule-based plan--action consistency on AndroidControl High~\citep{li2024androidcontrol} (+6.03\,pp), alongside improved visual grounding on ScreenSpot-v2 Mobile~\citep{wu2024osatlas}.

Three complementary diagnostics examine why these gains arise. Fixed-plan continuation isolates action generation, showing that plan-aligned action repair improves execution of a valid plan. Aligned-state retention analyses demonstrate that reference-aligned plan protection reduces regressions on baseline-correct actions. Free-generation comparisons further show that protecting plans preserves more valid plans and enables more joint repairs, supporting the complementary roles of protection and repair beyond surface plan--action agreement.

Our main contributions are threefold:
\begin{itemize}
    \item \textbf{Reference-aligned plan protection and plan-aligned action repair.} PACT uses reference-supported plans to decouple plan preservation from action correction without altering the scalar action reward.
    \item \textbf{Task improvements across mobile and desktop environments.} Comparisons against outcome-only and two repair baselines demonstrate consistent gains in dynamic task completion on AndroidWorld and OSWorld, supported by consistency and grounding evaluations on AndroidControl and ScreenSpot-v2 Mobile.
    \item \textbf{Mechanistic evidence on repair and retention.} Through fixed-plan continuation, free-generation, and baseline-correct retention analyses, we characterize the repair--retention trade-off while disentangling plan correctness, action validity, and plan--action consistency.
\end{itemize}

%% file: sec/fig1_execution_gap.tex
\begin{figure*}[t]
  \centering
  \begin{subfigure}[t]{0.41\textwidth}
    \centering
    \includegraphics[width=\linewidth]{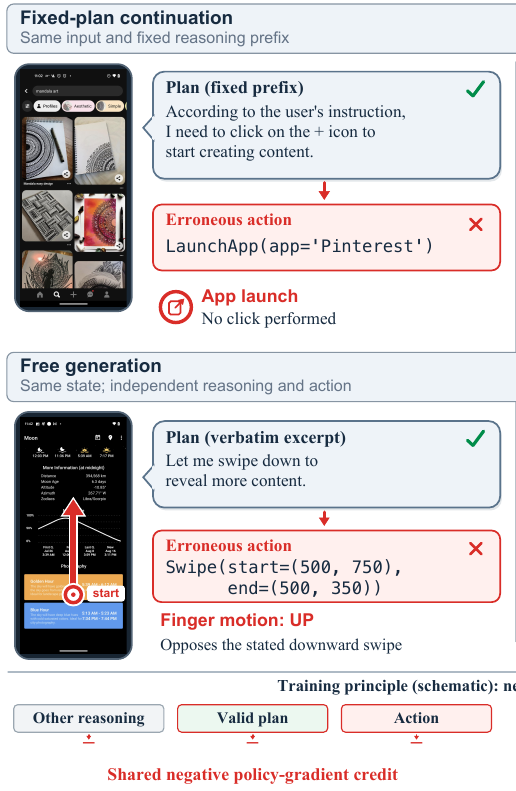}
    \caption{Original GRPO baseline (O)}
    \label{fig:execution-gap-outcome}
  \end{subfigure}%
  \begin{subfigure}[t]{0.41\textwidth}
    \centering
    \includegraphics[width=\linewidth]{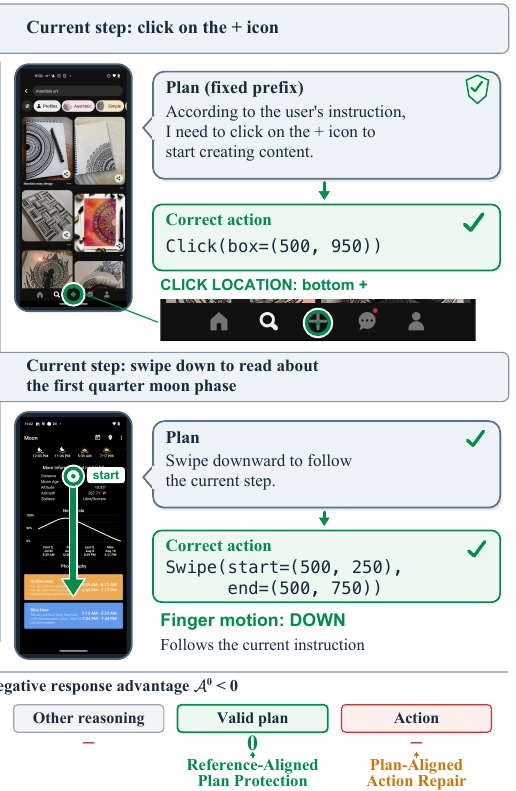}
    \caption{PACT}
    \label{fig:execution-gap-pact}
  \end{subfigure}
  \vspace{-2mm}
  \caption{\textbf{Execution gaps and targeted correction.}
  \textbf{Top:} Both models continue the same supplied Pinterest reasoning;
  O launches the app, while PACT clicks the intended \texttt{+} control.
  \textbf{Bottom:} With independently generated reasoning and actions on
  Moon, O states a downward swipe but emits upward finger motion, while
  PACT executes the downward swipe. On-screen circles mark the click or gesture
  start; swipe arrows show finger motion. The separate training schematic illustrates
  reference-aligned plan protection of selected plans and additional action repair
  under a negative response advantage.}
  \label{fig:execution-gap}
  \vspace{-3mm}
\end{figure*}

%% file: sec/2_related_work.tex
\section{Related Work}\label{related-work}

\textbf{Reasoning--execution alignment in GUI agents.}
Reliable GUI interaction demands both sound planning and faithful
execution, motivating fine-grained evaluation beyond aggregate task success.
VenusBench-Mobile~\citep{gong2026venusbench} decomposes user-centric
agent evaluation across perception, understanding, decision, action,
and memory capabilities, while Dong et al.~\citep{dong2025gap}
specifically diagnose reasoning--execution gaps by separating the
correctness of stated plans from that of emitted actions.
On the training side, Faithful-Agent~\citep{hu2026faithful} combines
evidence-grounded supervision, thought--action consistency rewards,
and a guided advantage estimator, whereas MobileIPL~\citep{huang2025mobileipl}
derives thinking-level preferences by propagating action feedback
through reasoning trees to improve planning.
In contrast, PACT uses reference-supported plans within
unsuccessful responses to determine the direction of correction---preserving
valid plans while repairing only their failed execution.

\textbf{Fine-grained preference learning and action correction.}
Building on the pairwise objective of DPO~\citep{rafailov2023dpo},
Step-DPO~\citep{lai2024stepdpo} targets individual reasoning steps
conditioned on a correct prefix, whereas Agentic-DPO~\citep{chen2026agenticdpo}
contrasts expert actions against sampled mistakes at expert states.
From a rollout-augmentation perspective, Octopus~\citep{ding2026octopus}
recombines sampled trajectories and masks specific segments to
separate self-correction from direct reasoning.
PACT instead retains the model's own decoded reasoning trace as the shared
conditioning context to contrast the reference-derived action suffix
against the emitted erroneous suffix.

\textbf{Credit assignment within failed responses.}
Localized supervision also raises a credit-assignment question: which
parts of an unsuccessful response should be penalized?
In standard GRPO~\citep{shao2024deepseekmath}, a single group-relative
advantage is broadcast across every token of a sampled response.
Recent methods refine this assignment based on intermediate
correctness, ranging from retrieval-calibrated negative advantages in
CalibAdv~\citep{wu2026calibadv} to teacher-identified first-error
boundaries in Cliff~\citep{han2026cliff}.
In GUI agent training, UI-Venus-2~\citep{venusteam_uivenus2} introduces
structured action-aware distillation to condition token-level
supervision on action correctness and distinguish action-type errors
from parameter errors.
Within outcome-supervised GRPO, PACT masks negative policy-gradient
contributions on reference-supported GUI plan spans and repairs
failed action suffixes under the same plan, while keeping the scalar
action reward intact.

%% file: sec/3_problem.tex
\section{Problem Formulation}\label{problem-formulation}

Let \(x\) denote the agent's observable input context, comprising a user
instruction, the current interface screenshot, and the available
interaction history. Given \(x\), the policy \(\pi_\theta\) autoregressively
generates a reasoning trace \(z\) followed by a GUI action \(a\):
\begin{equation}
\pi_\theta(z,a\mid x)
=\pi_\theta(z\mid x)\pi_\theta(a\mid x,z).
\label{eq:reasoning-action-factorization}
\end{equation}
During training, each context \(x\) is paired with an annotated reference
action \(a^\star\).

The \emph{plan span} in \(z\) expresses the next intended GUI
interaction. Building on reasoning--execution
diagnostics~\citep{dong2025gap}, Table~\ref{tab:quadrants} defines
\(T,A,C\) and joint repair, and summarizes the four correctness regimes
and their training implications in PACT.

All sampled responses remain in outcome-supervised GRPO.
PACT adds protection and repair for reference-supported plans whose
actions fail to implement them.
For incorrect plans, correction instead requires revising the reasoning;
valid actions should not be redirected to follow an incorrect plan.

\begin{table}[t]
\setlength{\tabcolsep}{2pt}
\centering\small
\caption{Step-level judgments and training implications in PACT.}
\label{tab:quadrants}
{\raggedright\textbf{(a) Definitions} (\(1\): satisfied; \(0\): otherwise)\par}
\vspace{2pt}
\begin{tabularx}{\linewidth}{@{}>{\centering\arraybackslash}p{49pt}>{\raggedright\arraybackslash}X@{}}
\toprule
Metric & Success criterion \\
\midrule
\(T\) & The current plan is correct for the task. \\
\(A\) & The action is valid for the current task and state. \\
\(C\) & The action implements the model's stated plan. \\
Joint repair & \(T=A=C=1\) for the same plan--action pair. \\
\bottomrule
\end{tabularx}
\par\vspace{5pt}
{\raggedright\textbf{(b) Correctness regimes}\par}
\vspace{2pt}
\begin{tabularx}{\linewidth}{@{}>{\centering\arraybackslash}p{12pt}>{\centering\arraybackslash}p{12pt}>{\raggedright\arraybackslash}X@{}}
\toprule
\(T\) & \(A\) & Training implication \\
\midrule
1 & 1 & Original action-reward objective. \\
1 & 0 & \textbf{Protect the plan; repair the action.} \\
0 & 1 & Original objective; no forced plan alignment. \\
0 & 0 & Original objective; no reward for error agreement. \\
\bottomrule
\end{tabularx}
\par\vspace{2pt}
\parbox{\linewidth}{\footnotesize Protection and repair apply only to
eligible gaps in Eq.~\eqref{eq:execution-gap-selector} (\(\widehat C_i=0\)).}
\end{table}

During training, a rule-based selector evaluates each sampled response
\(i\) relative to its reference action \(a_i^\star\). Let
\(J_i\in\{0,1\}\) indicate whether the response admits the required
reference-based judgments, and let
\(\widehat T_i,\widehat A_i,\widehat C_i\in\{0,1\}\) denote the
resulting proxies for \(T,A,C\). The execution-gap eligibility flag is
\begin{equation}
E_i = J_i\,\mathbf{1}\!\left[
\widehat T_i=1,\;\widehat A_i=0,\;\widehat C_i=0
\right].
\label{eq:execution-gap-selector}
\end{equation}
Unjudgeable responses (\(J_i=0\)) remain in the base RL objective without
auxiliary supervision. These reference-relative proxies govern
training-time selection, whereas benchmark evaluation and mechanism
diagnostics use independent rule-consistency and screenshot-based
semantic assessments.

%% file: sec/4_method.tex
\section{PACT}\label{er-r1}

Figure~\ref{fig:pact-framework} summarizes the original GRPO path and
PACT's two complementary training mechanisms.
\input{sec/fig2_spare_framework}

\subsection{Retaining the original action reward}
\label{retaining-the-original-action-learning-objective}

PACT retains the original scalar action reward without introducing any
plan-consistency bonus. For a sampled response \(y=(z,a)\), the
standard parseable branch of the mobile scorer combines a format score,
an action-type match indicator, and an action-specific parameter score:
\begin{equation}
R_0(y,a^\star)\!=\!r_{\mathrm{fmt}}(y)\!+\!\mathbf{1}[\tau(a)\!=\!\tau(a^\star)]
\bigl(1\!+\!P_0(a,a^\star)\bigr),
\label{eq:base-reward}
\end{equation}
where \(\tau(a)\) denotes the action type. All original parsing,
missing-action, and grounding branches remain unchanged, with
action-specific scoring rules and coordinate tolerances detailed in the
supplementary material.

For each prompt, the pre-update policy \(\pi_{\mathrm{old}}\) samples
\(K\) responses. Outcome-supervised GRPO~\citep{shao2024deepseekmath}
computes the group-standardized advantage
\begin{equation}
\mathcal A_i^0
=\frac{R_{0,i}-\mu_{g(i)}}{s_{g(i)}+10^{-6}},
\label{eq:grpo-advantage}
\end{equation}
where \(g(i)\) is the prompt group of response \(i\), and
\(\mu_{g(i)}\) and \(s_{g(i)}\) are its reward mean and sample standard
deviation. The baseline broadcasts \(\mathcal A_i^0\) to valid response
tokens in a clipped token-mean policy loss regularized by a KL term
against the fixed supervised fine-tuned (SFT) reference \(\pi_{\mathrm{ref}}\). Only sampled
rollouts enter these groups; repair suffixes never
affect group statistics (\(\mu_{g(i)}, s_{g(i)}\)) or base rollout rewards.

\subsection{Reference-verified execution-gap selection}
\label{selecting-a-verifiable-current-plan}

The selector evaluates the eligibility flag \(E_i\) in
Eq.~\eqref{eq:execution-gap-selector}. In the mobile implementation, it
scans the final four sentences of \(z_i\) in reverse order for an explicit
immediate plan, rejecting negation, multi-step ambiguity, and
conflicting action types. Candidate plans are extracted solely from
the reasoning text \(z_i\) prior to reference comparison, never inferred
retrospectively from the emitted action \(a_i\).

Judgeability (\(J_i=1\)) requires a parseable reference, a unique emitted action block, comparison
metadata, and a non-terminal plan with checkable parameters:
pointing actions require a unique explicit coordinate; scrolling actions
require a single direction; text-entry and app-launch actions require a
unique quoted argument; and supported parameter-free system actions
require a valid action type.
Grounding-only examples, terminal references, and unverifiable
plans (\(J_i=0\)) remain active in the base RL objective but are
excluded from auxiliary selection.

For judgeable responses, \(\widehat T_i\) checks the plan parameters
against \(a_i^\star\), \(\widehat A_i\) tests whether the base content
score reaches the configured full-credit threshold, and \(\widehat C_i\)
checks whether the emitted action \(a_i\) matches the plan under
distinct compatibility tolerances. When \(E_i=1\), span localization and
pair construction proceed independently, with extraction and fallback
rules in the supplement.

\subsection{Reference-aligned plan protection}
\label{protecting-only-the-selected-plan-span}

Let \(M^D_{i,t}\in\{0,1\}\) indicate tokens within the localized plan
span. PACT modifies the token-level policy advantage as
\begin{equation}
\widetilde{\mathcal A}_{i,t}=
\begin{cases}
\max(\mathcal A_i^0,0),& E_i=1\ \text{and}\ M^D_{i,t}=1,\\
\mathcal A_i^0,&\text{otherwise}.
\end{cases}
\label{eq:advantage-clamp}
\end{equation}
Without renormalizing group advantages, this clamp zeroes out negative
policy-gradient coefficients on the reference-supported plan span while
leaving positive advantages, action tokens, and the remaining reasoning
tokens untouched. Crucially, protection acts strictly on local policy-gradient
coefficients rather than freezing shared model parameters or masking the
KL regularizer.

\subsection{Plan-aligned action repair}
\label{repairing-the-action-under-the-same-plan}

For an eligible response \(y_i^-=(z_i,a_i^-)\), we construct a positive
response \(y_i^+=(z_i,a_i^+)\) by replacing only the action block with
the training reference \(a_i^\star\) formatted in the model's native
action syntax and coordinate space, preserving the input \(x_i\) and
decoded reasoning text \(z_i\). Let \(\mathcal P\) denote the retained
constructible pairs, where \(y_i^+\) is scored via teacher forcing;
unconstructible or overlength replacements are omitted from \(\mathcal P\),
while their original rollouts remain in the rollout batch \(\mathcal B\).

Let \(M^{A,\pm}_{i,t}\in\{0,1\}\) select the action tags and action
tokens, excluding reasoning and trailing tokens. We define the summed
action-span log probability as
\begin{equation}
s_{\theta,i}^{\pm}
=\sum_t M^{A,\pm}_{i,t}
\log\pi_\theta\bigl(y_{i,t}^{\pm}\mid x_i,y_{i,<t}^{\pm}\bigr),
\label{eq:action-logprob}
\end{equation}
so that only action-span tokens enter the auxiliary objective while the
decoded reasoning trace \(z_i\) serves as their shared conditioning context.

We evaluate the log-ratio margin using \(\pi_{\mathrm{old}}\) and
\(\pi_{\mathrm{ref}}\), and define the pair weight as
\begin{equation}
w_i=\frac{\lambda\beta}{B_0}
\sigma\!\bigl(\beta[s_{\mathrm{old},i}^{-}-s_{\mathrm{ref},i}^{-}
-s_{\mathrm{old},i}^{+}+s_{\mathrm{ref},i}^{+}]\bigr),
\label{eq:pair-weight}
\end{equation}
where \(\lambda\) is the repair weight, \(\beta\) is the preference scale,
\(\sigma\) is the sigmoid function, and \(B_0=|\mathcal B|\) is the
original rollout batch size (rather than \(|\mathcal P|\)). The
suffix-repair surrogate is
\begin{equation}
\mathcal L_{\mathrm{repair}}^{\mathrm{sur}}(\theta)
=\sum_{i\in\mathcal P}\operatorname{stopgrad}(w_i)
\bigl(s_{\theta,i}^{-}-s_{\theta,i}^{+}\bigr).
\label{eq:repair-surrogate}
\end{equation}
Minimizing Eq.~\eqref{eq:repair-surrogate} favors the reference-derived
suffix \(a_i^+\) over the erroneous suffix \(a_i^-\) under the same
decoded reasoning trace. With pairs and masks fixed, its gradient at
\(\theta_{\mathrm{old}}\) matches that of the corresponding conditional
logistic preference objective~\citep{rafailov2023dpo} when both suffix
terms are evaluated at that parameter point; reusing frozen weights
\(w_i\) across actor minibatches yields a linear log-probability
surrogate rather than a repeatedly recomputed logistic loss, with the
derivation and full rule specifications in the supplementary material.

Collecting all actor terms yields the complete objective:
\begin{equation}
\mathcal L_{\mathrm{actor}}(\theta)\!=\!
\mathcal L_{\mathrm{PG}}(\theta;\widetilde{\mathcal A},\mathcal B)
\!+\!\eta\mathcal L_{\mathrm{KL}}(\theta;\mathcal B)
\!+\!\mathcal L_{\mathrm{repair}}^{\mathrm{sur}}(\theta),
\label{eq:pact-objective}
\end{equation}
where \(\eta\) is the unchanged KL coefficient, and \(\lambda\) is absorbed into \(w_i\).
Constructed positive responses contribute exclusively to
\(\mathcal L_{\mathrm{repair}}^{\mathrm{sur}}\), without entering the rollout
batch \(\mathcal B\), the modified policy loss
\(\mathcal L_{\mathrm{PG}}(\theta;\widetilde{\mathcal A},\mathcal B)\), or the
KL term. At inference, PACT uses the standard single-pass
reasoning--action interface.

\subsection{Training configurations}
\label{four-training-configurations}

We compare four cumulative GRPO configurations in
Table~\ref{tab:training-configurations}.
In these configuration names, \textbf{O} denotes \textbf{O}riginal GRPO,
\textbf{A} denotes \textbf{A}ction Repair, and \textbf{R} denotes
\textbf{R}eference-Verified Execution-Gap Selection.

\begin{table}[t]
\centering
\footnotesize
\setlength{\tabcolsep}{2pt}
\renewcommand{\arraystretch}{1.05}
\captionsetup{skip=3pt}
\caption{Cumulative training configurations.}
\label{tab:training-configurations}
\renewcommand{\tabularxcolumn}[1]{m{#1}}
\begin{tabularx}{\columnwidth}{@{}>{\centering\arraybackslash}m{50pt}>{\centering\arraybackslash}m{35pt}*{2}{>{\centering\arraybackslash}X}@{}}
\toprule
Configuration & Action Repair & Reference-Verified Execution-Gap Selection & Reference-Aligned Plan Protection \\
\midrule
O & \(\times\) & \(\times\) & \(\times\) \\
OA & \(\checkmark\) & \(\times\) & \(\times\) \\
OAR & \(\checkmark\) & \(\checkmark\) & \(\times\) \\
PACT & \(\checkmark\) & \(\checkmark\) & \(\checkmark\) \\
\bottomrule
\end{tabularx}
\end{table}

\begin{itemize}
    \item \textbf{O (Original GRPO):} no action repair or plan protection.
    \item \textbf{OA (O + A):} adds action repair to O on supported, parseable, non-terminal actions below full credit (\(\widehat A_i=0\)), without execution-gap selection or plan protection.
    \item \textbf{OAR (O + A + R):} adds reference-verified execution-gap selection (\(E_i=1\)) to OA, yielding plan-aligned action repair without advantage clamping (\(\widetilde{\mathcal A}_{i,t}=\mathcal A_i^0\)).
    \item \textbf{PACT:} adds reference-aligned plan protection to OAR.
\end{itemize}
All variants retain the original reward and reference KL. Repair variants
construct pairs from their own rollouts using the same reference-action
constructor and suffix surrogate. Optimization settings are in
Section~\ref{experimental-setup} and the supplement.

%% file: sec/fig2_spare_framework.tex
\begin{figure*}[t]
  \centering
  \captionsetup{skip=0pt}
  \includegraphics[width=\textwidth,trim=0 12bp 0 0,clip]{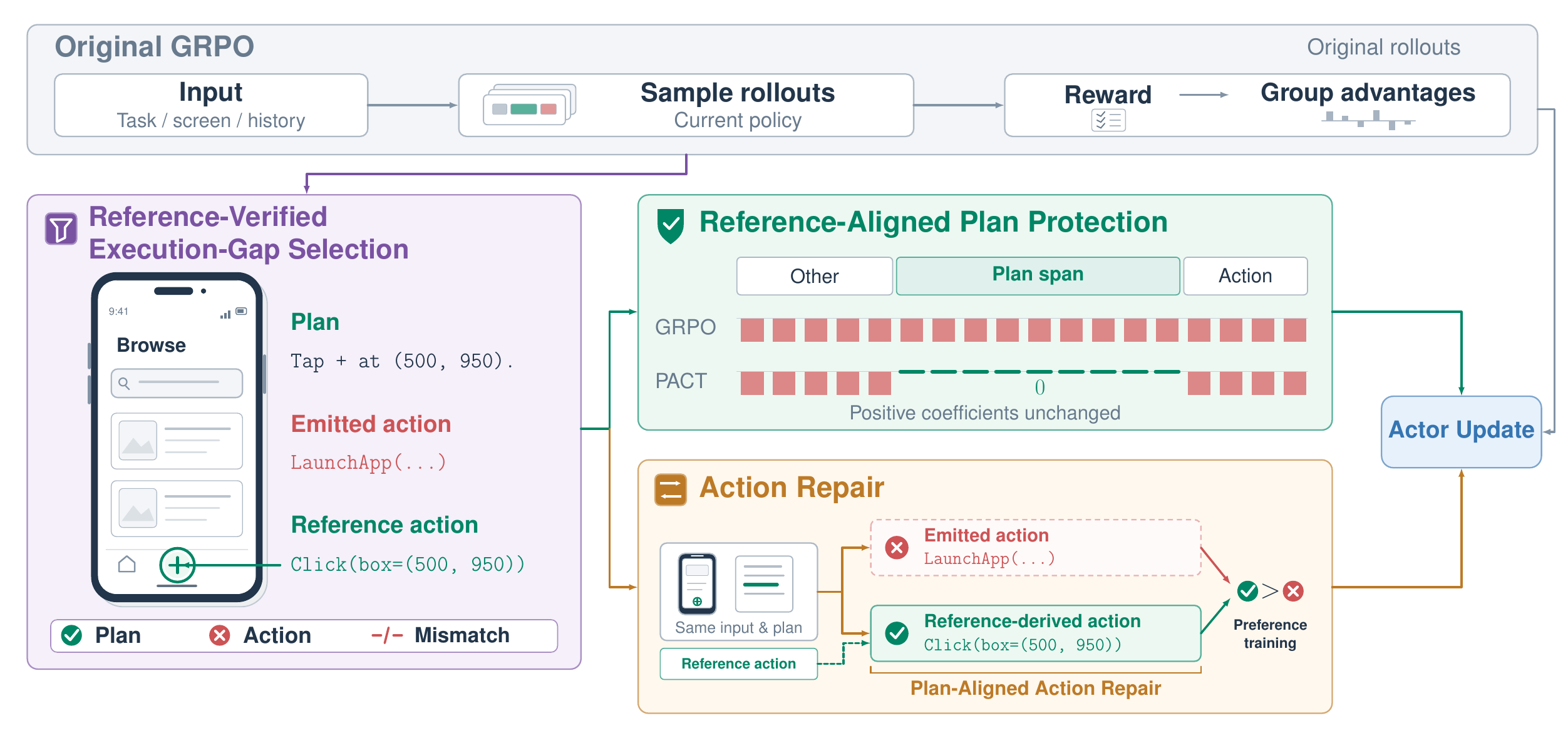}
  \vspace{-3mm}
  \caption{PACT training framework. Reference-verified execution-gap selection identifies reference-supported plans
  paired with incorrect, inconsistent actions. Reference-aligned plan protection
  zeros negative policy-gradient coefficients on the plan span. Plan-aligned
  action repair favors reference-derived over erroneous action suffixes under
  the same input and reasoning trace. Both feed the GRPO actor update.}
  \label{fig:pact-framework}
  \vspace{-3mm}
\end{figure*}

%% file: sec/5_experiments.tex
\section{Experimental Setup}\label{experimental-setup}

\input{tables/pact_main_table}

\textbf{Training.}
In our primary mobile study, O, OA, OAR, and PACT share the same
Qwen3.5-9B vision--language model SFT initialization, whose frozen copy
serves as the reference policy \(\pi_{\mathrm{ref}}\), along with 10,000
mobile GUI state--action training examples with pixel-space reference coordinates.
Each variant is trained for 1,250 training iterations over two epochs,
sampling 16 prompts and \(K=8\) responses per prompt at each step to
yield \(B_0=128\) original rollout responses before auxiliary rows are
added. Optimization uses one PPO epoch per rollout batch, a learning
rate of \(10^{-6}\), rollout temperature \(1.0\), KL coefficient
\(\eta=0.01\), and 32 accelerators with FSDP2; repair-enabled variants
set \(\lambda=\beta=0.1\) and retain at most 64 repair pairs per
iteration. To evaluate cross-platform generality, we also train
desktop-adapted counterparts of O, OA, OAR, and PACT. Platform-specific
training settings, action-space adaptations, and optimization details
are provided in the supplementary material.

\textbf{Benchmarks and roles.}
Our evaluation separates interactive task completion, plan--action
consistency, and visual grounding.
On AndroidWorld~\citep{rawles2024androidworld}, we report the task success
rate across its 116 interactive tasks for each checkpoint.
On AndroidControl~\citep{li2024androidcontrol}, we evaluate 8,444 states
from 1,543 episodes in each of the High and Low settings: along with the
current screenshot and available history, High provides only
the task goal, whereas Low additionally supplies the step instruction.
In both settings, we measure rule-based plan--action consistency
and reference-action matching for retention and mechanism
diagnostics.
ScreenSpot-v2 Mobile~\citep{wu2024osatlas} measures point-in-target hit
rate across 501 examples (290 text and 211 icon targets).
For desktop interaction, we report task success for each
desktop checkpoint on the full 369-task OSWorld
benchmark~\citep{xie2024osworld}, counting a task as successful
strictly when its native score equals \(1.0\).

\textbf{Generation and measurement.}
Mobile evaluation uses greedy decoding (temperature \(0\)), a
4,096-token output limit, and up to ten prior interaction steps. On
AndroidControl, models observe recorded screenshots, with non-initial
states conditioned on each model's own generated history. OSWorld uses
the same environment configuration, greedy decoding (temperature \(0\)),
a 100-step limit, and the official native evaluator.
Distinct from the training-time selector, our AndroidControl consistency
evaluator extracts a function-style or keyword-based action intent from
the reasoning trace and compares the inferred action type and recoverable
parameters with those of the emitted action. We report
\(C_{\mathrm{rule}}=N_{\mathrm{consistent}}/N_{\mathrm{judgeable}}\)
with consistent and judgeable counts provided in the supplementary material.
For action retention, we measure the proportion of O's reference-matched
states on which a compared variant also matches the reference.

In mechanism diagnostics, fixed-plan continuation conditions each model
on the same input and the exact valid reasoning prefix generated by O to isolate action
generation, whereas free generation requires each model to produce both
reasoning and action from scratch. On these subsets,
screenshot-based semantic assessments use language-model assistance
and human verification to separately evaluate plan correctness (\(T\)),
action validity (\(A\)), and semantic plan--action consistency (\(C\)),
complementing benchmark-wide rule-based scores; the supplementary
material details sample selection, annotation protocols, and execution.

%% file: tables/pact_main_table.tex
\begin{table*}[t]
\centering
\definecolor{pacttableink}{HTML}{243449}
\definecolor{pacttablewash}{HTML}{EDF6F1}
\definecolor{pactcontextink}{HTML}{586775}
\newcommand{\PACTpaneltitle}[2]{\par\begingroup\raggedright\fontsize{9}{10.5}\selectfont\color{#1}\bfseries #2\par\endgroup\vspace{1pt}}
\newcommand{\PACTth}[1]{{\fontsize{9}{10.5}\selectfont\strut #1}}
\newcommand{\PACTtablenote}[1]{\par\vspace{1pt}\begingroup\raggedright\fontsize{8}{9.3}\selectfont #1\par\endgroup}
\captionsetup{skip=3pt}
\caption{\textbf{Main benchmark results.} (A) Published GUI-agent context. (B) Matched RL comparisons.}
\label{tab:pact-comprehensive}
\setlength{\tabcolsep}{4pt}
\renewcommand{\arraystretch}{1.0}
\setlength{\aboverulesep}{1.5pt}
\setlength{\belowrulesep}{1.5pt}

{\fontsize{10}{11.2}\selectfont
\PACTpaneltitle{pactcontextink}{(A) Published benchmark results: representative 7B--9B GUI agents}
\begin{tabularx}{\linewidth}{>{\raggedright\arraybackslash}p{127pt}>{\centering\arraybackslash}p{23pt}*{3}{>{\centering\arraybackslash}X}}
\toprule
\PACTth{Model} & \PACTth{Size} & \PACTth{\mbox{AndroidWorld SR (\%)}} & \PACTth{\mbox{OSWorld SR (\%)}} & \PACTth{ScreenSpot-v2 Mobile (\%)} \\
\midrule
UI-TARS-7B-SFT~\citep{qin2025uitars} & 7B & 33.0 & 17.7 & 93.61\,\citep{qin2025uitars} \\
UI-TARS-1.5-7B~\citep{uitars15card} & 7B & 30.0\,\citep{zhou2025maiui} & 27.5 & 91.23\,\citep{zhou2025guiaima} \\
UI-Venus-Navi-7B~\citep{gu2025uivenus} & 7B & 49.1 & -- & -- \\
GUI-Owl-7B~\citep{ye2025guiowl} & 7B & 66.4 & 29.5\,\citep{chen2026step} & -- \\
Qwen3-VL-8B-Instruct~\citep{zhou2025maiui} & 8B & 47.6 & -- & 92.38\,\citep{zhou2025maiui} \\
Step-GUI-8B~\citep{yan2025stepgui} & 8B & 67.7 & 40.2 & -- \\
Qwen3.5-9B~\citep{qwen2026qwen35card} & 9B & 57.8 & 41.8 & 93.01\,\citep{groundingpi2026} \\
\midrule
\textbf{PACT (ours)} & 9B & \textbf{68.97} & \textbf{44.17} & \textbf{96.01} \\
\bottomrule
\end{tabularx}
\PACTtablenote{SR: task success rate. ``--'': not listed. Published-result settings and aggregation details are provided in the supplement.}
}

\vspace{3pt}
{\small
\PACTpaneltitle{pacttableink}{(B) Main benchmark results under matched settings within each platform}
\begin{tabularx}{\linewidth}{>{\raggedright\arraybackslash}p{100pt}*{4}{>{\centering\arraybackslash}X}>{\centering\arraybackslash}p{84pt}}
\toprule
\multirow{2}{*}{Configuration} & \multicolumn{2}{c}{Task success $\uparrow$} & \multicolumn{2}{c}{Plan--action consistency $\uparrow$} & Grounding $\uparrow$ \\
\cmidrule(lr){2-3}\cmidrule(lr){4-5}\cmidrule(l){6-6}
 & AndroidWorld & OSWorld & AC High $C_{\mathrm{rule}}$ & AC Low $C_{\mathrm{rule}}$ & ScreenSpot-v2 Mobile \\
\midrule
O & 60.34 & 40.92 & 83.71 & 86.39 & 91.82 \\
OA & 61.21 & 41.19 & 80.04 & 78.47 & \textbf{96.01} \\
OAR & 62.93 & 42.55 & 86.43 & \textbf{88.14} & 95.41 \\
\midrule
\rowcolor{pacttablewash}
\textbf{PACT} & \textbf{68.97} & \textbf{44.17} & \textbf{89.74} & 87.56 & \textbf{96.01} \\
\midrule
$\Delta$(PACT$-$O), pp & +8.63 & +3.25 & +6.03 & +1.17 & +4.19 \\
\bottomrule
\end{tabularx}
\PACTtablenote{AC: AndroidControl; $C_{\mathrm{rule}}$: rule consistency over judgeable states. Scores: \%; $\Delta$: pp.}
}
\end{table*}

%% file: sec/6_results.tex
\section{Results and Analysis}\label{results-and-analysis}

\subsection{Mobile task completion, consistency, and grounding}
\label{task-completion-consistency-and-grounding}

\textbf{PACT improves interactive task completion.}
On AndroidWorld, PACT leads the matched RL configurations in
Table~\ref{tab:pact-comprehensive}(B) at \textbf{68.97\%} task success:
\textbf{+8.63\,pp} over O (60.34\%), +6.04\,pp over OAR, and +7.76\,pp
over OA. Table~\ref{tab:pact-comprehensive}(A) provides published context.
In a file-copy task, PACT selects the target folder, confirms copying, and
finishes in 15 steps. OAR repeatedly types the folder name until
the step limit (Supplement Figure~S2).

\textbf{Task gains are accompanied by stronger plan--action consistency.}
On AndroidControl High, PACT leads in \(C_{\mathrm{rule}}\) at
\textbf{89.74\%} (+6.03\,pp over O, +3.31\,pp over OAR, +9.70\,pp over OA),
with the highest consistent-action count (7,046 states) and
41.1\% fewer action-type mismatches than O, from 1,279 to 753
(Supplement Table~S3). On Low, PACT raises \(C_{\mathrm{rule}}\) over O
from 86.39\% to 87.56\%, while OAR reaches 88.14\% but incurs 217 more
action regressions than PACT
(Section~\ref{repairing-failures-while-retaining-baseline-correct-actions}).

\textbf{Visual grounding improves alongside task performance.}
On ScreenSpot-v2 Mobile, PACT raises the point-in-target hit rate from
91.82\% (460/501) to \textbf{96.01\%} (481/501), improving on both text
and icon targets and exceeding OAR (95.41\%;
Table~\ref{tab:pact-comprehensive}(B)). PACT has the highest grounding score in
Table~\ref{tab:pact-comprehensive}(A).
Notably, while PACT and OA tie
on aggregate grounding accuracy, PACT leads OA by 7.76\,pp on \mbox{AndroidWorld},
\mbox{indicating} that PACT's advantage in multi-step task completion
extends beyond static point localization.

\subsection{Extending PACT to desktop GUI agents}
\label{extending-to-desktop-agents}

\textbf{The task-level gains extend to desktop interaction.}
On the full 369-task OSWorld benchmark (Table~\ref{tab:pact-comprehensive}(A,B)), PACT solves
\textbf{163 tasks (44.17\%)}, surpassing O (151, 40.92\%) by
\textbf{12 tasks}, OAR (157, 42.55\%) by \textbf{6 tasks}, and
OA (152, 41.19\%) by 11 tasks. Action repair adds one success over O,
while OAR and PACT yield larger gains, matching AndroidWorld's
\(\mathrm{PACT}>\mathrm{OAR}>\mathrm{OA}>O\) ordering. On LibreOffice CSV export, PACT selects
\texttt{Text CSV} and finishes in 9 steps. OAR misclicks \texttt{PNG}
and reaches the 100-step limit (Supplement Figure~S3).

\subsection{Repairing failures while retaining baseline-correct actions}
\label{repairing-failures-while-retaining-baseline-correct-actions}

\begin{figure*}[t]
\centering
\includegraphics[width=0.90\textwidth]{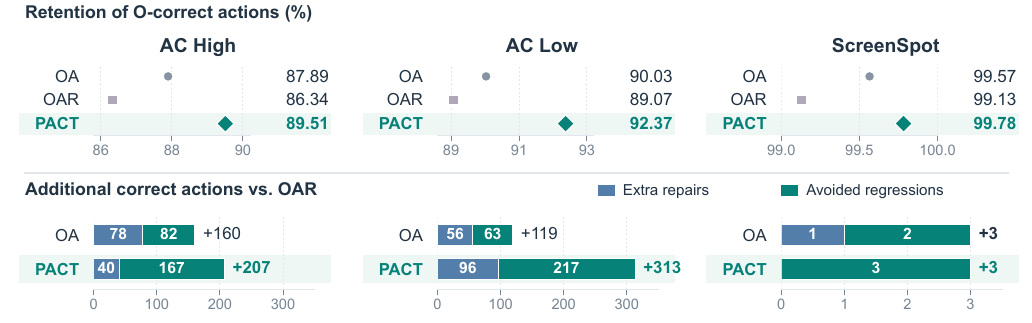}
\vspace{-2mm}
\caption{Retention and repair on aligned states (higher is better). Top: O-correct action retention. Bottom: net correct-action gains over OAR, computed as extra repairs plus avoided regressions. PACT leads retention across settings and net gains on AndroidControl.}\label{fig-retention}
\end{figure*}

Effective repair must correct errors while retaining mastered behavior.
Let \(\mathcal K_0\) denote states where O emits a reference-matching
action. Table~\ref{tab:retention} reports retention, recovered errors,
regressions, and net error change \(\Delta E\) relative to O;
Figure~\ref{fig-retention} shows net correct-action gains over OAR.

\begin{table}[t]
\setlength{\tabcolsep}{2pt}
\centering\small
\captionsetup{skip=3pt}
\caption{Reference-action retention on aligned AndroidControl states. \(|\mathcal K_0|\): High 5,270; Low 6,580. Non-initial states use each model's own generated history. W/R: wrong/right. Net error change vs.\ O: \(\Delta E=(R{\to}W)-(W{\to}R)\).}
\begin{tabularx}{\linewidth}{@{}>{\centering\arraybackslash}p{35pt}>{\centering\arraybackslash}p{25pt}>{\centering\arraybackslash}p{35pt}>{\centering\arraybackslash}p{32pt}>{\centering\arraybackslash}p{32pt}>{\centering\arraybackslash}X@{}}
\toprule
Setting & Model & Ret. (\%) & W\(\to\)R \(\uparrow\) & R\(\to\)W \(\downarrow\) & Net \(\Delta E\downarrow\) \\
\midrule
\multirow{3}{*}{AC High} & OA & 87.89 & 669 & 638 & \(-31\) \\
 & OAR & 86.34 & 591 & 720 & \(+129\) \\
 & PACT & \textbf{89.51} & 631 & 553 & \(\mathbf{-78}\) \\
\midrule
\multirow{3}{*}{AC Low} & OA & 90.03 & 436 & 656 & \(+220\) \\
 & OAR & 89.07 & 380 & 719 & \(+339\) \\
 & PACT & \textbf{92.37} & 476 & 502 & \(\mathbf{+26}\) \\
\bottomrule
\end{tabularx}
\label{tab:retention}
\end{table}

\textbf{Plan protection improves both retention and repair relative to OAR.}
PACT achieves the highest baseline-correct retention on AndroidControl at \textbf{89.51\% on High and 92.37\% on Low}, versus
86.34\% and 89.07\% for OAR, preserving 167 and 217 additional
baseline-correct states (23.2\% and 30.2\% fewer regressions) while
recovering 40 and 96 more baseline errors, totaling \textbf{+207} and
\textbf{+313} correct actions over OAR on High and Low, compared with
+160 and +119 for OA (Figure~\ref{fig-retention}).

\textbf{Limiting regressions yields a superior trade-off over action repair.}
On AC High, OA recovers 38 more baseline errors than PACT (669 vs.\ 631),
but PACT avoids 85 more regressions (553 vs.\ 638), yielding \textbf{47 more
reference-action matches overall}. On Low, PACT leads on both axes, preserving
154 more baseline-correct states and recovering 40 more errors for a
\textbf{net advantage of 194 matches}. Thus, curbing regressions on
mastered states translates directly into higher overall reference action accuracy
than either repair baseline.

A similar retention advantage holds on ScreenSpot-v2 Mobile: PACT preserves
\textbf{459/460 (99.78\%)} of O-correct targets (vs.\ 456/460 for OAR and
458/460 for OA), while all three variants acquire 22--23 new hits.

\subsection{Fixed-plan execution versus free generation}
\label{fixed-plan-execution-versus-free-generation}

We next examine execution on 23 identified AndroidControl Low initial
states with empty history, where O stated a valid immediate plan but
emitted an incompatible application-launch action
(Table~\ref{tab:fixed-plan}). Under fixed-plan continuation, all four
models—including O—generate a fresh action continuation from the same
screenshot, instruction, and exact O reasoning prefix, whereas free
generation requires each model to produce both reasoning and action.

\begin{table}[t]
\setlength{\tabcolsep}{2pt}
\centering\small
\caption{Execution diagnostics on the 23-state execution-gap set. Asterisks (*) mark O's selection criteria. Joint repair is defined in Table~\ref{tab:quadrants}(a). Full semantic results are in the supplementary material.}
\begin{tabularx}{\linewidth}{@{}>{\raggedright\arraybackslash}p{108pt}>{\centering\arraybackslash}X>{\centering\arraybackslash}X>{\centering\arraybackslash}X>{\centering\arraybackslash}X@{}}
\toprule
Metric (\%; out of 23 states) & O & OA & OAR & PACT \\
\midrule
\multicolumn{5}{@{}l}{\textbf{Fixed plan}} \\
\hspace{3pt}Reference action accuracy & 56.52 & 95.65 & \textbf{100.0} & \textbf{100.0} \\
\addlinespace[2pt]
\multicolumn{5}{@{}l}{\textbf{Free generation}} \\
\hspace{3pt}Reference action accuracy & 0.0* & \textbf{95.65} & \textbf{95.65} & \textbf{95.65} \\
\hspace{3pt}Plan--action consistency (\(C\)) & 0.0* & \textbf{100.0} & \textbf{100.0} & \textbf{100.0} \\
\hspace{3pt}Semantic joint repair & 0.0* & \textbf{100.0} & 95.65 & 95.65 \\
\bottomrule
\end{tabularx}
\label{tab:fixed-plan}
\end{table}

\textbf{Plan-aligned action repair resolves fixed-plan execution failures.}
Conditioned on the fixed valid plan, both PACT and OAR raise reference action
accuracy from \textbf{56.52\% (13/23)} for O to \textbf{100.0\% (23/23)},
while OA reaches 95.65\% (22/23); Supplement Figure~S4 shows cross-app
examples. Holding the input and plan fixed isolates action generation,
showing that suffix supervision translates a valid plan into
the intended action.

\textbf{The repaired behavior transfers to free generation.}
When generating both plans and actions autonomously, OA, OAR, and PACT all
achieve 95.65\% (22/23) reference action accuracy; under semantic
evaluation, PACT and OAR achieve 95.65\% (22/23) joint repair while OA reaches
100.0\% (23/23). The additional semantic success for OA comes from a valid
off-center product click outside the reference point-distance tolerance,
as detailed in the supplementary material. Although all three
reach 100.0\% (23/23) plan--action consistency (\(C\)), their
joint-repair scores separate surface agreement with a generated plan
from faithful execution of the requested step.

\subsection{Preserving plans during free generation}
\label{complementary-natural-gap-cases-and-failure-boundaries}

To assess reference-aligned plan protection, we compare PACT with OAR in Table~\ref{tab:natural-gaps}
on 30 natural execution-gap states covering action-type,
scroll-direction, and input-order errors, with protocol details in the supplementary material.

\begin{table}[t]
\setlength{\tabcolsep}{1pt}
\centering\small
\caption{Free-generation results on all 30 states and the 24 with \(C_{\mathrm{OAR}}=C_{\mathrm{PACT}}=1\), using Table~\ref{tab:quadrants}(a) metrics.}
\begin{tabularx}{\linewidth}{@{}>{\raggedright\arraybackslash}p{85pt}>{\centering\arraybackslash}p{25pt}*{3}{>{\centering\arraybackslash}X}>{\centering\arraybackslash}p{46pt}@{}}
\toprule
Subset & Model & \(T\uparrow\) & \(A\uparrow\) & \(C\uparrow\) & Joint repair\(\uparrow\) \\
\midrule
\multirow{2}{*}{All (\(N=30\))}
 & OAR & 19/30 & 13/30 & 25/30 & 13/30 \\
 & PACT & \textbf{23/30} & \textbf{19/30} & \textbf{28/30} & \textbf{19/30} \\
\midrule
\multirow{2}{*}{\shortstack[l]{{\footnotesize Both models consistent}\\\(C=1\), \(N=24\)}}
 & OAR & 13/24 & 12/24 & \textbf{24/24} & 12/24 \\
 & PACT & \textbf{18/24} & \textbf{16/24} & \textbf{24/24} & \textbf{16/24} \\
\bottomrule
\end{tabularx}
\label{tab:natural-gaps}
\end{table}

\Needspace{3\baselineskip}
\textbf{PACT preserves more valid plans and completes more repairs.}
PACT leads OAR on all four metrics, raising joint repair from 43.33\% to
\textbf{63.33\% (+20.00\,pp)}. Among the 24 states where both models
are plan--action consistent, PACT preserves five more correct plans
(\textbf{18/24} vs.\ 13/24) and completes four more joint repairs
(\textbf{16/24} vs.\ 12/24). For example, in Supplement Figure~S1, PACT closes the blocking Myntra
popup before typing, whereas OAR types into the covered field; Figures~S5--S8 illustrate additional single-step and boundary cases.

%% file: sec/7_limitations.tex
\section{Discussion and Limitations}\label{discussion-and-limitations}

\textbf{Why plan protection complements action repair.}
Across multi-step benchmarks, the complete PACT framework achieves
substantially higher task success on AndroidWorld and the highest task
success on OSWorld within the matched comparisons of
Table~\ref{tab:pact-comprehensive}(B),
while preserving more baseline-correct actions across aligned states
in Table~\ref{tab:retention}, even though OAR and OA match or approach PACT on
isolated single-step grounding in Table~\ref{tab:pact-comprehensive}(B)
and fixed-plan continuation in Table~\ref{tab:fixed-plan}. These results clarify
the complementary roles of our two mechanisms: plan-aligned action repair teaches
the policy to translate a valid plan into the intended action,
whereas reference-aligned plan protection shields that plan from negative
policy-gradient updates on failed rollouts. Because each action in
closed-loop interaction alters the context for subsequent turns,
retaining mastered behaviors is as critical as recovering from errors,
enabling PACT to drive stronger end-to-end task completion across mobile
and desktop environments.

\textbf{Limitations and future directions.}
First, our training-time selector requires explicit plans whose
parameters can be verified against a single annotated reference.
Consequently, valid but implicit plans or semantically equivalent
alternative actions may fall outside the eligible set; incorporating
semantic plan verifiers or multi-reference supervision could
broaden the coverage of repairable failures.
Second, whereas our OSWorld experiments establish task-level gains on
desktop workflows, our fine-grained plan and fixed-plan
continuation diagnostics focus on mobile states. Extending these
mechanistic analyses to richer desktop action spaces—such as keyboard
shortcuts, composite operations, and long-horizon workflows—will
clarify how both mechanisms interact in complex environments.

%% file: sec/8_conclusion.tex
\section{Conclusion}\label{conclusion}

We presented \textbf{PACT}, a training framework that couples
\emph{reference-aligned plan protection} with \emph{plan-aligned action repair}
to resolve GUI execution gaps without altering the original scalar
action reward. It shields reference-supported plan spans from
negative policy-gradient updates while optimizing corrected action
suffixes under the same decoded reasoning trace. Compared with outcome-only
RL, PACT substantially improves AndroidWorld task success and achieves
higher task success on OSWorld, alongside mobile plan--action consistency
and visual grounding. Aligned-state retention analyses further show
that PACT preserves a higher proportion of baseline-correct actions
than unprotected repair variants. Moreover, fixed-plan continuation and
free-generation diagnostics confirm the complementary roles of both
mechanisms while distinguishing true action correctness from
surface-level plan--action agreement. Together, these findings
establish a correctness-guided credit-assignment principle for GUI
agents: preserve valid plans and selectively repair their
execution, rather than optimizing consistency alone.

%% file: sec/X_suppl.tex
\section{Implementation Details}\label{implementation-details}

Sections~\ref{original-action-reward}--\ref{actor-integration-and-optimization-schedule}
detail the reward, selection and construction rules, gradient derivation,
and optimization schedule. Section~\ref{evaluation-definitions-and-additional-tables}
provides benchmark breakdowns and diagnostic protocols, and
Section~\ref{qualitative-cases} presents qualitative cases.

\subsection{Original action reward and coordinate tolerances}\label{original-action-reward}

Within each platform, all four training configurations share the scalar
outcome scorer \(R_0(y,a^\star)\) defined in Eq.~(3) of the main paper.
In the mobile scorer's standard parseable branch, the format term \(r_{\mathrm{fmt}}(y)\) assigns \(+0.5\) for valid
\texttt{<think>}...\texttt{</think><action>}...\texttt{</action>}
structure and \(-0.5\) otherwise. Matching the reference action type
\(\tau(a)=\tau(a^\star)\) contributes \(1.0\) point, followed by the
action-specific parameter score \(P_0(a,a^\star)\). Missing actions,
unparseable outputs, and grounding-only samples retain the base scorer's
original branches.

For mobile point-based actions (e.g., \texttt{Click}, \texttt{LongPress}),
predicted coordinates are emitted on a normalized \(0\text{--}999\) grid
and mapped back to native image pixels. Let \(d\) denote the Euclidean
pixel distance between the predicted and reference coordinates, and let
\(m=\min(W,H)\) be the shorter screen dimension. The parameter score is
\begin{equation}
P_{\mathrm{click}}(d)=\begin{cases}
1.0,& d<0.07m,\\
0.5,& 0.07m\leq d<0.14m,\\
0.0,& d\geq 0.14m.
\end{cases}
\label{eq:supp-click-reward}
\end{equation}
For \texttt{Scroll}/\texttt{Swipe} actions with directional metadata, the
scorer checks both gesture endpoints and assigns \(1.5\) parameter points
when both satisfy the strict coordinate tolerance. The branch with
endpoint coordinates alone uses its standard Boolean endpoint comparison.
Text-based actions (\texttt{Type}, \texttt{LaunchApp}, \texttt{Answer})
receive full parameter credit under exact normalized string match or a
token-set F1 score exceeding \(0.5\).

\subsection{Reference-verified execution-gap selection and plan span localization}\label{reference-relative-selector}

To compute the execution-gap eligibility flag
\(E_i=J_i\,\mathbf{1}[\widehat T_i=1,\widehat A_i=0,\widehat C_i=0]\) in
Eq.~(2) of the main paper, the programmatic selector first evaluates
judgeability \(J_i\in\{0,1\}\). Setting \(J_i=1\) requires a unique
emitted action block, a parseable reference action \(a_i^\star\) with
comparison metadata, and a non-terminal plan with checkable
parameters. Grounding-only samples and terminal references
(\texttt{Finished}, \texttt{Answer}, \texttt{CallUser}) are assigned
\(J_i=0\) and participate in the base GRPO objective.

The following rules define the three binary proxies for judgeable mobile
responses (\(J_i=1\)).
\begin{itemize}
    \item \textbf{Action correctness proxy (\(\widehat A_i\)):}
    \(\widehat A_i=1\) if the base content score (action-type bonus plus
    parameter score) reaches the configured threshold.
    The selector uses a threshold of \(2.5\) for \texttt{Scroll}/\texttt{Swipe}
    and \(2.0\) for other standard actions. Responses below the
    threshold receive \(\widehat A_i=0\).

    \item \textbf{Plan correctness proxy (\(\widehat T_i\)):}
    The reasoning trace \(z_i\) is segmented into clauses using line
    breaks and English/Chinese punctuation, and the final four sentences
    are scanned in reverse order. A valid plan must specify a
    single immediate action without negation or multi-step ambiguity.
    Setting \(\widehat T_i=1\) requires the extracted plan
    parameters to match \(a_i^\star\). Pointing plans require a
    unique explicit \(0\text{--}999\) coordinate within the strict
    \(0.07\min(W,H)\) pixel tolerance of \(a_i^\star\). Directional
    plans require a unique matching direction. Text and app-launch
    plans require a unique quoted argument matching the normalized
    reference string. Parameter-free system actions must match by action
    type.

    \item \textbf{Plan--action compatibility proxy (\(\widehat C_i\)):}
    \(\widehat C_i=1\) if the emitted action \(a_i\) matches the
    extracted plan in action type, lies within a grid distance of
    \(1\) on the normalized \(0\text{--}999\) coordinate space for
    pointing actions, and matches any specified direction, text, or
    application argument.
\end{itemize}

For pointing actions, \(\widehat T_i\) checks proximity to the reference
coordinate, whereas \(\widehat C_i\) checks agreement with the explicitly
stated coordinate. PACT selects repair candidates using the full
eligibility condition \(E_i=1\), including \(\widehat A_i=0\).

For reference-aligned plan protection (\(E_i=1\)), the character span of the
verified plan is mapped to response token indices to form the binary
mask \(M^D_{i,t}\in\{0,1\}\). The implementation uses tokenizer character
offsets whenever the re-encoded reasoning prefix matches the active token
sequence, and otherwise performs an exact token-subsequence search prior
to the action tag. If a unique token span cannot be localized, we set
\(M^D_{i,t}=0\) for all \(t\), leaving the base advantage
\(\mathcal A_i^0\) unchanged while still permitting suffix repair if pair
construction succeeds.

\subsection{Constructing positive action suffixes}\label{constructing-positive-action-suffixes}

For each eligible mobile response \(y_i^-=(z_i,a_i^-)\), the positive action
suffix \(a_i^+\) is synthesized from the training reference \(a_i^\star\)
in the model's native action syntax and \(0\text{--}999\) coordinate grid.
For an eligible response, the selector has verified the extracted plan
against the annotated reference action. We use that action to construct
the positive suffix under the original reasoning trace.
Reference pixel coordinates are converted to the normalized grid, rounded,
and clamped to valid bounds. Scroll references with endpoint coordinates
are formatted as \texttt{Swipe} actions. Reference actions with missing
endpoints or text arguments that cannot be serialized in the native
action syntax, as well as overlength replacements, are omitted from
\(\mathcal P\). Their original rollouts remain in \(\mathcal B\).

On both platforms, the constructed positive sequence
\(y_i^+=(z_i,a_i^+)\) retains the exact
input context \(x_i\) (instruction, screenshot, and interaction history)
and the decoded reasoning text \(z_i\), replacing only the
\texttt{<action>}...\texttt{</action>} block. The binary action masks
\(M^{A,-}_{i,t}\) and \(M^{A,+}_{i,t}\) cover only the action tags and
action tokens, excluding all reasoning tokens and any trailing text. At
most 64 constructible pairs are retained per training iteration in batch
order. In the OA baseline (original GRPO with Action Repair), pairs are
constructed using the identical reference formatter from parseable, non-terminal rollouts
with \(\widehat A_i=0\) and \(a_i^-\neq a_i^+\), without requiring
\(\widehat T_i=1\) or \(\widehat C_i=0\).

\subsection{First-order gradient derivation of the suffix-repair surrogate}\label{local-relation-of-the-surrogate-to-logistic-preference-optimization}

Let \(s_{\theta,i}^\pm\) denote the summed action-span log probability
defined in Eq.~(6) of the main paper, and let the log-ratio margin be
\begin{equation}
\ell_i(\theta)
=\bigl(s_{\theta,i}^+-s_{\mathrm{ref},i}^+\bigr)
-\bigl(s_{\theta,i}^--s_{\mathrm{ref},i}^-\bigr).
\label{eq:supp-margin}
\end{equation}
A conditional DPO-style logistic preference objective~\citep{rafailov2023dpo,lai2024stepdpo}
over the retained pair set \(\mathcal P\) is given by
\begin{equation}
\mathcal L_{\mathrm{log}}(\theta)
=-\frac{\lambda}{B_0}\sum_{i\in\mathcal P}
\log\sigma\bigl(\beta\ell_i(\theta)\bigr).
\label{eq:supp-dpo-loss}
\end{equation}
Here, \(\lambda\) is the repair weight and \(\beta\) is the preference scale.
Taking the gradient with respect to \(\theta\) and evaluating at the
pre-update policy \(\theta=\theta_{\mathrm{old}}\) (where
\(\ell_i(\theta_{\mathrm{old}})=\ell_i^{\mathrm{old}}\)) yields
\begin{equation}
\begin{aligned}
&\left.\nabla_\theta\mathcal L_{\mathrm{log}}(\theta)\right|_{\theta=\theta_{\mathrm{old}}}\\
&\quad=\sum_{i\in\mathcal P}\underbrace{\frac{\lambda\beta}{B_0}
\sigma\bigl(-\beta\ell_i^{\mathrm{old}}\bigr)}_{w_i}
\left.\bigl(\nabla_\theta s_{\theta,i}^- -\nabla_\theta s_{\theta,i}^+\bigr)\right|_{\theta=\theta_{\mathrm{old}}}.
\end{aligned}
\label{eq:supp-dpo-grad}
\end{equation}
Detaching the pre-update weight \(w_i\) (where
\(-\ell_i^{\mathrm{old}}=s_{\mathrm{old},i}^- -s_{\mathrm{ref},i}^-
-s_{\mathrm{old},i}^+ +s_{\mathrm{ref},i}^+\) matches Eq.~(7) of the main paper)
via \(\operatorname{stopgrad}(w_i)\)
produces the linear suffix-repair surrogate
\(\mathcal L_{\mathrm{repair}}^{\mathrm{sur}}(\theta)
=\sum_{i\in\mathcal P}\operatorname{stopgrad}(w_i)(s_{\theta,i}^- -s_{\theta,i}^+)\)
in Eq.~(8) of the main paper, whose gradient matches Eq.~\eqref{eq:supp-dpo-grad}
at \(\theta=\theta_{\mathrm{old}}\).

\subsection{Optimization schedule, hyperparameters, and desktop adaptation}\label{actor-integration-and-optimization-schedule}

\begin{table*}[t]
\centering\small
\caption{Training hyperparameters for the mobile and desktop studies. Repair parameters apply to OA, OAR, and PACT.}\label{tab:supp-hyperparams}
\begin{tabularx}{\textwidth}{@{}>{\raggedright\arraybackslash}X>{\centering\arraybackslash}p{0.25\textwidth}>{\centering\arraybackslash}p{0.25\textwidth}@{}}
\toprule
Setting & Mobile & Desktop \\
\midrule
Initialization and fixed reference & Qwen3.5-9B SFT & Qwen3.5-9B SFT \\
RL training examples & 10,000 & 18,000 \\
Training iterations / epochs & 1,250 / 2 & 562 / 1 \\
Prompts per iteration / responses per prompt (\(K\)) & 16 / 8 & 32 / 8 \\
Original rollout batch size (\(B_0\)) & 128 & 256 \\
Optimizer & AdamW & Distributed Adam \\
Learning rate / PPO epochs & \(10^{-6}\) / 1 & \(10^{-6}\) / 1 \\
Learning-rate schedule & Constant & Constant \\
Weight decay / gradient clipping & 0.01 / 1.0 & 0.01 / 1.0 \\
Policy-loss aggregation / entropy coefficient & Token mean / 0 & Token mean / 0 \\
Probability-ratio clipping / dual clipping & 0.2 / 3.0 & 0.2 / 3.0 \\
Sampling temperature / top-\(p\) / top-\(k\) & 1.0 / 1.0 / \(-1\) & 1.0 / 1.0 / \(-1\) \\
Maximum prompt / response tokens & 36,864 / 2,048 & 40,960 / 2,048 \\
Accelerators / training framework & 32 / FSDP2 & 64 / Megatron \\
Rollout tensor parallelism & 4 & 4 \\
Numerical precision & bfloat16 & bfloat16 \\
KL coefficient (\(\eta\)) / estimator & 0.01 / \texttt{k3+} & 0.01 / \texttt{low\_var\_kl} \\
Repair weight (\(\lambda\)) / preference scale (\(\beta\)) & 0.1 / 0.1 & 0.1 / 0.1 \\
Maximum repair pairs per iteration & 64 & 64 \\
\bottomrule
\end{tabularx}
\end{table*}

\textbf{Training data and optimization.}
Table~\ref{tab:supp-hyperparams} summarizes the RL configurations
for both platforms. All four configurations (O, OA, OAR, and PACT;
main Table~2) on both platforms start RL training from the same SFT checkpoint.
We obtain this checkpoint by supervised fine-tuning the
Qwen3.5-9B vision--language model on multimodal GUI trajectories
to generate structured
\texttt{<think>...\allowbreak</think>\allowbreak<action>...\allowbreak</action>}
responses, with spatial action coordinates normalized to
\(0\text{--}999\). A frozen copy of this checkpoint serves as
the KL reference policy \(\pi_{\mathrm{ref}}\).
Within each platform, the four variants share the base-training
configuration in Table~\ref{tab:supp-hyperparams}.
Desktop RL training uses 18,000 examples sampled without
replacement (seed 42) and runs
for one epoch with complete 32-prompt batches (562 iterations), preserving
the original system instructions, interaction history,
screenshots, and reference actions. All training data are
strictly disjoint from the evaluation benchmarks.

The desktop action interface supports pointing, keyboard, drag/scroll,
and control actions. Training references are dominated by
\texttt{Click} (43.0\%), \texttt{Hotkey} (26.1\%), and
\texttt{Type} (21.2\%). Each training example pairs the multimodal input
context (system instructions, interaction history, and screenshots)
with its ground-truth reference action.

Both platforms use token-mean policy loss and apply KL regularization
in the actor objective with \(\eta=0.01\). The mobile \texttt{k3+}
estimator combines a clipped \(k_3\) value with a straight-through
squared-log-ratio gradient. Desktop training uses the \texttt{low\_var\_kl} estimator
\(e^d-d-1\), with
\(d=\log\pi_{\mathrm{ref}}-\log\pi_\theta\) clipped to \([-20,20]\)
and the estimator clipped to \([-10,10]\).
Original rollout groups determine GRPO advantages. Constructed positive
responses carry the suffix-repair loss with zero PG and KL masks, and
signed pair weights \(\pm w_i\) preserve the sum in
Eq.~(8) of the main paper under data-parallel averaging.
On 32 accelerators, the mobile 1,250-iteration training loop takes
\(120.76\) hours for O and \(123.19\) hours for PACT, excluding initial
model loading, corresponding to approximately \(2.0\%\) overhead.

Algorithm~\ref{alg:supp-pact} summarizes one mobile PACT training
iteration. Selection, span localization, and pair construction use
Sections~\ref{reference-relative-selector} and~\ref{constructing-positive-action-suffixes}.

\begin{algorithm}[t]
\caption{One mobile PACT training iteration}\label{alg:supp-pact}
\footnotesize
\begin{algorithmic}[1]
\Require Pre-update policy \(\pi_{\mathrm{old}}\), reference \(\pi_{\mathrm{ref}}\), 16 prompts with reference actions, \(K=8\), pair cap 64
\State Sample original rollout batch \(\mathcal B\), \(B_0=128\)
\State Compute \(R_0\) and grouped \(\mathcal A_i^0\) using main Eqs.~(3)--(4)
\State Broadcast \(\widetilde{\mathcal A}_{i,t}\gets\mathcal A_i^0\); set \(\mathcal P\gets\varnothing\)
\For{each response \(i\) in batch order}
  \State Compute \(J_i,\widehat T_i,\widehat A_i,\widehat C_i\) and \(E_i\) using main Eq.~(2)
  \If{\(E_i=1\)}
    \State Localize \(M^D_{i,t}\) (use a zero mask if no unique span is found)
    \State Clamp plan-token advantages using main Eq.~(5)
    \State Independently construct \(y_i^+=(z_i,a_i^+)\) from \(a_i^\star\)
    \If{the pair is constructible and \(|\mathcal P|<64\)}
      \State Append the pair and its action masks to \(\mathcal P\)
    \EndIf
  \EndIf
\EndFor
\If{\(\mathcal P\ne\varnothing\)}
  \State Pack positive responses into a 128-row auxiliary batch with zero-loss padding
  \State Score both suffixes under \(\pi_{\mathrm{old}}\) and \(\pi_{\mathrm{ref}}\); compute and freeze \(w_i\) using main Eqs.~(6)--(7)
  \State Assign \(+w_i\) to sampled negative suffixes, \(-w_i\) to constructed positive suffixes, and zero to padding
  \State Set auxiliary PG and KL masks to zero
\EndIf
\State Update policy parameters \(\theta\) by minimizing the actor objective in main Eq.~(9)
\end{algorithmic}
\end{algorithm}

\textbf{Desktop plan verification and suffix construction.}
The desktop selector applies the same execution-gap condition to
explicit plans. Table~\ref{tab:supp-desktop-actions}
sets out the supported action parameters. Reference support
\(\widehat T_i\) compares the plan with the reference action,
while compatibility \(\widehat C_i\) compares the emitted action with
the plan. Both comparisons require matching action types.
Desktop action correctness \(\widehat A_i=1\) requires the full native
content score of \(2.0\), comprising one point for type and one for
parameters.

\begin{table*}[t]
\centering\footnotesize
\caption{Desktop training-action verification and reference-derived suffix construction. Spatial distances are Euclidean distances on the normalized \(0\text{--}999\) grid.}\label{tab:supp-desktop-actions}
\begin{tabularx}{\textwidth}{@{}>{\raggedright\arraybackslash}p{64pt}>{\raggedright\arraybackslash}p{80pt}>{\raggedright\arraybackslash}X>{\raggedright\arraybackslash}p{105pt}@{}}
\toprule
Action & Required parameters & Reference support / plan compatibility & Positive suffix construction \\
\midrule
\texttt{Click}, \texttt{DoubleClick}, \texttt{RightClick} & One \texttt{box} coordinate & \(\widehat T\): distance \(<49.95\) (\(0.05\times999\)). \(\widehat C\): distance \(\leq1\). & Copy reference \texttt{box}. \\
\texttt{Drag}, \texttt{Swipe} & \texttt{start} and \texttt{end} coordinates & \(\widehat T\): each endpoint distance \(<139.86\) (\(0.14\times999\)). \(\widehat C\): each endpoint distance \(\leq1\). & Copy both reference endpoints. \\
\texttt{Type} & One quoted \texttt{content} string & \(\widehat T,\widehat C\): exact string equality, preserving whitespace and escapes. & Copy reference \texttt{content}. \\
\texttt{Hotkey} & One nonempty \texttt{keys} list or tuple & \(\widehat T,\widehat C\): equality of normalized, ordered key tuples. & Copy reference \texttt{keys}. \\
\texttt{Wait} & No arguments & \(\widehat T,\widehat C\): matching \texttt{Wait} action. & \texttt{Wait()}. \\
\midrule
\texttt{CallUser}, \texttt{Finished} & Optional \texttt{content} string & Terminal actions receive native scoring with exact content matching. & Excluded from auxiliary selection. \\
\bottomrule
\end{tabularx}
\end{table*}

For desktop pointing actions, the native parameter score is \(1.0\)
at grid distances below \(49.95=0.05\times999\), \(0.5\) from
\(49.95\) up to but excluding \(99.9=0.10\times999\), and zero otherwise.
The reference-support test uses the
full-credit threshold. Drag and Swipe plans specify exactly two
coordinate pairs in start--end order. For \texttt{Type}, the native
parameter score awards \(1.0\) to exact text matches and \(0.5\) to
non-exact matches with the same trailing-whitespace category and
token-set F1 greater than \(0.5\). \texttt{Hotkey} comparison strips
surrounding whitespace and lowercases each key while preserving tuple
order and the literal space key.

Desktop reference coordinates already use the model's normalized grid.
The positive suffix serializes the reference action name and parameters
with Python literal representations and is checked by the same action
parser. The resulting pair retains the original input and decoded
reasoning context, following Section~\ref{constructing-positive-action-suffixes}.

\section{Evaluation Protocols and Additional Quantitative Results}\label{evaluation-definitions-and-additional-tables}

\subsection{AndroidControl rule-based consistency}\label{rule-consistency-and-coverage}

On AndroidControl, the rule-based consistency evaluator first searches
the reasoning trace for the final function-style action expression; if
none is present, it extracts the last English or Chinese action keyword
after resolving keyword overlaps, and parses any accompanying coordinate,
direction, or text arguments. For pointing actions with coordinates in both
the extracted plan and the emitted action, a coordinate mismatch is counted
when their pixel distance exceeds $0.14\sqrt{W^2 + H^2}$.
Disagreements between the extracted intent
and the emitted action are categorized into action-type or direction mismatches
(\(N_{\mathrm{type}}\)), coordinate mismatches (\(N_{\mathrm{coord}}\)),
and text-argument mismatches (\(N_{\mathrm{text}}\)). Rule-based
consistency is computed over all judgeable outputs as
\begin{equation}
C_{\mathrm{rule}}
=\frac{N_{\mathrm{consistent}}}{N_{\mathrm{consistent}}+N_{\mathrm{type}}+N_{\mathrm{coord}}+N_{\mathrm{text}}}.
\label{eq:supp-c-rule}
\end{equation}
Responses lacking a reasoning block, an extractable action intent, or a
valid reference label are excluded from the denominator of
\(C_{\mathrm{rule}}\). Table~\ref{tab:supp-ac-coverage} reports consistent
and judgeable counts over 8,444 states per setting.
These states comprise all annotated action steps in the 1,543 evaluation
episodes, with identical episode--step pairs used for High and Low.
On AC High, PACT achieves the highest \(C_{\mathrm{rule}}\) (89.74\%) and
absolute consistent-action count (7,046 vs.\ O: 6,879; OAR: 6,997; OA: 6,440),
remaining first with all 8,444 states as a common denominator.
Action-type mismatches account for most rule inconsistencies. PACT reduces
these mismatches from 1,279 to 753 on High, and has the fewest coordinate
mismatches on both High (16) and Low (6).

\begin{table*}[t]
\centering\small
\captionsetup{skip=3pt}
\setlength{\tabcolsep}{4pt}
\caption{AndroidControl consistency and mismatch counts over 8,444 states per setting. Type, coordinate, and text columns partition the inconsistent judgeable outputs. Bold marks the best values within each setting, including ties.}\label{tab:supp-ac-coverage}
\begin{tabular*}{\textwidth}{@{\extracolsep{\fill}}llrrrrr@{}}
\toprule
Setting & Model & Consistent / judgeable & Type\(\downarrow\) & Coord.\(\downarrow\) & Text\(\downarrow\) & \(C_{\mathrm{rule}}\)(\%)\(\uparrow\) \\
\midrule
AC High & O & 6879 / 8218 & 1279 & 19 & 41 & 83.71 \\
 & OA & 6440 / 8046 & 1559 & 25 & \textbf{22} & 80.04 \\
 & OAR & 6997 / 8096 & 1032 & 21 & 46 & 86.43 \\
 & PACT & \textbf{7046} / 7852 & \textbf{753} & \textbf{16} & 37 & \textbf{89.74} \\
\midrule
AC Low & O & 7182 / 8313 & 1081 & 19 & 31 & 86.39 \\
 & OA & 6534 / 8327 & 1753 & 24 & \textbf{16} & 78.47 \\
 & OAR & \textbf{7270} / 8248 & \textbf{906} & 24 & 48 & \textbf{88.14} \\
 & PACT & 7155 / 8172 & 966 & \textbf{6} & 45 & 87.56 \\
\bottomrule
\end{tabular*}
\vspace{-9pt}
\end{table*}

Because \(C_{\mathrm{rule}}\) relies on surface keyword and function
extraction, it provides a scalable benchmark-wide proxy rather than a
full semantic parse. For example, mentioning an in-app ``Home'' tab can
trigger the system \texttt{PressHome} keyword rule, and discussing a
subsequent step at the end of a multi-clause sentence can override an
earlier immediate click intent. These boundary cases motivate our
complementary screenshot-based semantic diagnostics in the main paper.

\subsection{Interactive task evaluation on AndroidWorld and OSWorld}\label{androidworld-aggregation}

\textbf{AndroidWorld.}
For each mobile checkpoint, we report the task success rate across
AndroidWorld's 116 interactive tasks, yielding 70/116 (60.34\%) for O,
71/116 (61.21\%) for OA, 73/116 (62.93\%) for OAR,
and \textbf{80/116 (68.97\%)} for PACT.

\textbf{OSWorld.}
All four desktop-adapted variants (O, OA, OAR, and PACT) are evaluated on the full 369-task
OSWorld benchmark. Evaluation uses a standardized Linux
x86/Docker environment with a read-only Ubuntu base image and the official
native evaluation scripts. All four variants use temperature \(0\)
and a 100-step interaction limit. Input construction
and generation parameters are given in Section~\ref{prompt-and-generation-protocols}.
At each step, the complete emitted action block is parsed and executed.
A task is counted as successful
strictly when its native evaluator returns a full score of \(1.0\):
\begin{equation}
\mathrm{SR}_{\mathrm{OSWorld}}
=\frac{1}{369}\sum_{j=1}^{369}\mathbf{1}\bigl[\mathrm{native\_score}_j=1.0\bigr],
\label{eq:supp-osworld-sr}
\end{equation}
without counting partial completion scores as successes.
The totals are 151/369 (40.92\%) for O, 152/369 (41.19\%) for OA,
157/369 (42.55\%) for OAR, and \textbf{163/369 (44.17\%)} for PACT.
PACT completes 12 more tasks than O, 11 more than OA, and 6 more than OAR.
Main Table~3(A) gives external baselines under the evaluation settings of
the cited papers and model cards, while Table~3(B) compares our four RL
configurations within each platform.
The Step-GUI-8B and Qwen3.5-9B scores in Table~3(A) use OSWorld-Verified.

\subsection{ScreenSpot-v2 Mobile subgroup breakdown and target retention}\label{grounding-and-retention}

Tables~\ref{tab:supp-screenspot-subgroups} and~\ref{tab:supp-screenspot-retention}
provide the target-category breakdown and baseline-correct retention
statistics on ScreenSpot-v2 Mobile. Because ScreenSpot-v2 Mobile consists
of single-step grounding queries without multi-turn history, Table~\ref{tab:supp-screenspot-retention}
offers a history-independent test of capability preservation: PACT
retains \(99.78\%\) (\(459/460\)) of O-correct targets while incurring
only a single right-to-wrong regression.
For the four cited ScreenSpot-v2 Mobile scores in main Table~3(A),
text/icon accuracies are weighted by 290/211 targets.

\begin{table*}[t]
\centering\small
\captionsetup{skip=3pt}
\caption{ScreenSpot-v2 Mobile point-in-target accuracy across all 501 examples, 290 text targets, and 211 icon targets.}\label{tab:supp-screenspot-subgroups}
\begin{tabularx}{\textwidth}{@{}>{\raggedright\arraybackslash}X>{\centering\arraybackslash}X>{\centering\arraybackslash}X>{\centering\arraybackslash}X@{}}
\toprule
Configuration & All targets (501) & Text targets (290) & Icon targets (211) \\
\midrule
O & 460 (91.82\%) & 271 (93.45\%) & 189 (89.57\%) \\
OA & \textbf{481 (96.01\%)} & \textbf{286 (98.62\%)} & 195 (92.42\%) \\
OAR & 478 (95.41\%) & \textbf{286 (98.62\%)} & 192 (91.00\%) \\
PACT & \textbf{481 (96.01\%)} & 285 (98.28\%) & \textbf{196 (92.89\%)} \\
\bottomrule
\end{tabularx}
\vspace{-9pt}
\end{table*}

\begin{table*}[t]
\centering\small
\captionsetup{skip=3pt}
\caption{Target-level retention and repair on ScreenSpot-v2 Mobile (460 O-correct targets).}\label{tab:supp-screenspot-retention}
\begin{tabularx}{\textwidth}{@{}>{\raggedright\arraybackslash}X>{\centering\arraybackslash}X>{\centering\arraybackslash}X>{\centering\arraybackslash}X>{\centering\arraybackslash}X@{}}
\toprule
Configuration & Retained / 460 & Retention (\%) & Wrong to right (\(\uparrow\)) & Right to wrong (\(\downarrow\)) \\
\midrule
OA & 458 / 460 & 99.57 & \textbf{23} & 2 \\
OAR & 456 / 460 & 99.13 & 22 & 4 \\
PACT & \textbf{459 / 460} & \textbf{99.78} & 22 & \textbf{1} \\
\bottomrule
\end{tabularx}
\vspace{-10pt}
\end{table*}

\begin{table*}[!t]
\setlength{\tabcolsep}{4pt}
\centering\small
\captionsetup{skip=3pt}
\caption{Complete fixed-plan continuation and free-generation results on the 23-state execution-gap diagnostic set (\%, with exact counts out of 23 in parentheses). Joint repair requires \(T=A=C=1\). Asterisks (*) mark O's selection criteria. Bold indicates row-wise maxima, including ties.}
\begin{tabular}{@{}lcccc@{}}
\toprule
Condition / metric & O & OA & OAR & PACT \\
\midrule
Fixed plan: reference action accuracy & 56.52 (13/23) & 95.65 (22/23) & \textbf{100.0 (23/23)} & \textbf{100.0 (23/23)} \\
Free generation: reference action accuracy & 0.0 (0/23)* & \textbf{95.65 (22/23)} & \textbf{95.65 (22/23)} & \textbf{95.65 (22/23)} \\
Free generation: plan correctness (\(T\)) & \textbf{100.0 (23/23)}* & \textbf{100.0 (23/23)} & 95.65 (22/23) & 95.65 (22/23) \\
Free generation: action validity (\(A\)) & 0.0 (0/23)* & \textbf{100.0 (23/23)} & 95.65 (22/23) & 95.65 (22/23) \\
Free generation: plan--action consistency (\(C\)) & 0.0 (0/23)* & \textbf{100.0 (23/23)} & \textbf{100.0 (23/23)} & \textbf{100.0 (23/23)} \\
Free generation: semantic joint repair & 0.0 (0/23)* & \textbf{100.0 (23/23)} & 95.65 (22/23) & 95.65 (22/23) \\
\bottomrule
\end{tabular}
\label{tab:supp-fixed-plan-full}
\vspace{-10pt}
\end{table*}

\begin{table*}[t!]
\centering\small
\captionsetup{skip=3pt}
\caption{Natural execution-gap composition and free-generation results on all 30 states, the 24 with \(C_{\mathrm{OAR}}=C_{\mathrm{PACT}}=1\), and the remaining six. Bold marks the better value within each comparison, including ties.}\label{tab:supp-natural-gaps}
\begin{minipage}[t]{0.30\textwidth}
\textbf{(a) Set composition}\par\vspace{3pt}
\footnotesize
\setlength{\tabcolsep}{1.25pt}
\begin{tabular*}{\linewidth}{@{}>{\raggedright\arraybackslash}p{0.87\linewidth}@{\extracolsep{\fill}}r@{}}
\toprule
Candidate category & States \\
\midrule
Click-versus-launch conflict & 8 \\
Scroll-direction error & 2 \\
Premature text entry / skipped selection & 17 \\
Click-versus-wait conflict & 1 \\
Swipe-versus-click/selection conflict & 1 \\
Ambiguous-reference boundary case & 1 \\
\midrule
Total & 30 \\
\bottomrule
\end{tabular*}
\end{minipage}\hfill
\begin{minipage}[t]{0.68\textwidth}
\textbf{(b) Free-generation results (\%, with exact counts)}\par\vspace{3pt}
\footnotesize
\setlength{\tabcolsep}{1.5pt}
\begin{tabular*}{\linewidth}{@{}l@{\extracolsep{\fill}}ccccc@{}}
\toprule
Subset & Model & \(T\uparrow\) & \(A\uparrow\) & \(C\uparrow\) & Joint repair\(\uparrow\) \\
\midrule
\multirow{2}{*}{All (\(N=30\))}
 & OAR & 63.33 (19/30) & 43.33 (13/30) & 83.33 (25/30) & 43.33 (13/30) \\
 & PACT & \textbf{76.67 (23/30)} & \textbf{63.33 (19/30)} & \textbf{93.33 (28/30)} & \textbf{63.33 (19/30)} \\
\midrule
\multirow{2}{*}{\shortstack[l]{Both models consistent\\\(C=1\), \(N=24\)}}
 & OAR & 54.17 (13/24) & 50.00 (12/24) & \textbf{100.0 (24/24)} & 50.00 (12/24) \\
 & PACT & \textbf{75.00 (18/24)} & \textbf{66.67 (16/24)} & \textbf{100.0 (24/24)} & \textbf{66.67 (16/24)} \\
\midrule
\multirow{2}{*}{Remaining (\(N=6\))}
 & OAR & \textbf{100.0 (6/6)} & 16.67 (1/6) & 16.67 (1/6) & 16.67 (1/6) \\
 & PACT & 83.33 (5/6) & \textbf{50.00 (3/6)} & \textbf{66.67 (4/6)} & \textbf{50.00 (3/6)} \\
\bottomrule
\end{tabular*}
\end{minipage}
\vspace{-10pt}
\end{table*}

\subsection{Mechanism diagnostic protocols and complete results}\label{conditional-mechanism-protocols}

Language models assisted only with initial candidate screening for both
sets. The authors checked each plan and action against the task and
screenshot, including click locations, and assigned all final labels.
\(T\) measures plan correctness for the task, \(A\) action validity in
the current state and task, and \(C\) whether the action implements
the model's stated plan. Joint repair requires
\(T=A=C=1\) for the same plan--action pair, as in Table~1(a) of the main paper.
When a valid plan (\(T=1\)) is paired with an invalid action (\(A=0\)),
consistency is judged as \(C=0\) if the action targets a different UI
element or operation than stated (e.g., landing on an unrelated control
in Figure~S7), and \(C=1\) if it targets the stated element or text but
fails task validity due to boundary localization or interface state constraints.

\textbf{23-state fixed-plan continuation and free generation.}
The 23 diagnostic states were identified among AndroidControl Low initial
steps (empty interaction history) where O articulated a valid immediate plan
but emitted an erroneous application-launch action. Under the
\emph{fixed-plan} condition, each model receives the identical task
instruction, screenshot, empty history, and O's exact valid reasoning
prefix terminating at the \texttt{<action>} opening tag, and deterministically
generates the action continuation in a single turn.
For a matched comparison with OA, OAR, and PACT, O also generates a fresh
action continuation from the same fixed prefix, yielding \(13/23\);
its \(0/23\) free-generation score comes from the original outputs
used to select these states.
Under \emph{free generation}, the assistant prefix is
removed and each model generates both reasoning and action from scratch.
Reference-action matching and screenshot-based semantic evaluation are
reported side by side in Table~5 of the main paper. On the single state
where OA's semantic action-validity and reference-matching judgments differ
(a Nike product-card screen), OA clicks a valid off-center region of the
product image outside the point-distance tolerance of the single reference
coordinate, whereas PACT and OAR score \(22/23\) under both criteria.
Table~\ref{tab:supp-fixed-plan-full} reports the complete fixed-plan
continuation and free-generation results, including the semantic breakdown.
Following our coordinate-based action interface, reference-action
matching evaluates predictions directly against each state's annotated
reference action \(a^\star\) (without accessibility-tree bounding boxes or
cross-type equivalences), requiring matching action types and
reference-compatible parameters (\(0.14\sqrt{W^2+H^2}\) point-distance
tolerance for clicks, matching swipe direction, and normalized text match).

\textbf{30-state natural execution-gap set.}
We selected candidates from O's AndroidControl High and Low responses
before collecting their paired OAR and PACT outputs, targeting valid plans
paired with invalid, inconsistent actions. Eleven states use empty history
and 19 use O's interaction history. OAR and PACT freely generate plans and
actions from identical screenshots, instructions, and available O
histories, without O's current response or the reference action.
This set and the 23-state fixed-plan diagnostic follow distinct
selection protocols and are analyzed separately.

All 30 states remain in each denominator; only verified successes enter
the numerator. The ambiguous-reference boundary case scores as a
non-success for \(T\) and joint repair, with \(C\) evaluated independently.
In Table~\ref{tab:supp-natural-gaps}, PACT raises \(T\) from 63.33\% (19/30)
to \textbf{76.67\% (23/30)}, \(C\) from 83.33\% (25/30) to
\textbf{93.33\% (28/30)}, and both \(A\) and joint repair from
43.33\% (13/30) to \textbf{63.33\% (19/30, +20.00\,pp)}.
Among the 24 states with \(C_{\mathrm{OAR}}=C_{\mathrm{PACT}}=1\), PACT retains \textbf{18/24}
correct plans versus 13/24 for OAR (+20.83\,pp) and completes
\textbf{16/24} joint repairs versus 12/24 (+16.67\,pp).
On the remaining six states, OAR generates six valid plans but only one
consistent action and joint repair; PACT achieves consistency on four
states and joint repair on three.

\input{sec/supp_prompt_protocols}

\clearpage
\onecolumn
\section{Qualitative Case Studies}\label{qualitative-cases}
\subsection{Comparative Success Cases Across Mobile and Desktop}\label{comparative-success-cases}

\textbf{Preserving the next-step requirement.}
The Myntra example in Figure~\ref{fig-supp-myntra} uses the same screenshot
and two preceding O-history steps for PACT and OAR. The task requires closing
the modal before text entry. PACT preserves that ordering in its immediate
plan and clicks the close button. OAR assumes that the covered search
field is ready and emits \texttt{Type}. Both outputs have \(C=1\), while
only PACT meets the plan, action, and joint-repair criteria. This example
illustrates how PACT preserves a valid plan and completes the repair during
free generation.

\textbf{Executing a complete mobile workflow.}
In Figure~\ref{fig-supp-aw-receipt}, all four configurations receive the
same receipt-copy goal and task seed. PACT selects Download, confirms the
copy, and emits \texttt{Finished} at step 15. OAR emits
\texttt{Type(content='Download')} at steps 9, 10, 12, 14, and 16.
The native final score is \(1.0\) for PACT and \(0.0\) for O, OA, and OAR.

\textbf{Completing a desktop export task.}
Figure~\ref{fig-supp-os-csv} shows the same OSWorld task across the four
configurations. PACT uses the Save As workflow, selects Text CSV, confirms
the export options, and finishes in nine steps. Its exported file passes
the official CSV-content comparison. O, OA, and OAR receive native scores
of zero at the 100-step limit.

\noindent\begin{minipage}{\textwidth}
\centering
\includegraphics[width=0.84\textwidth]{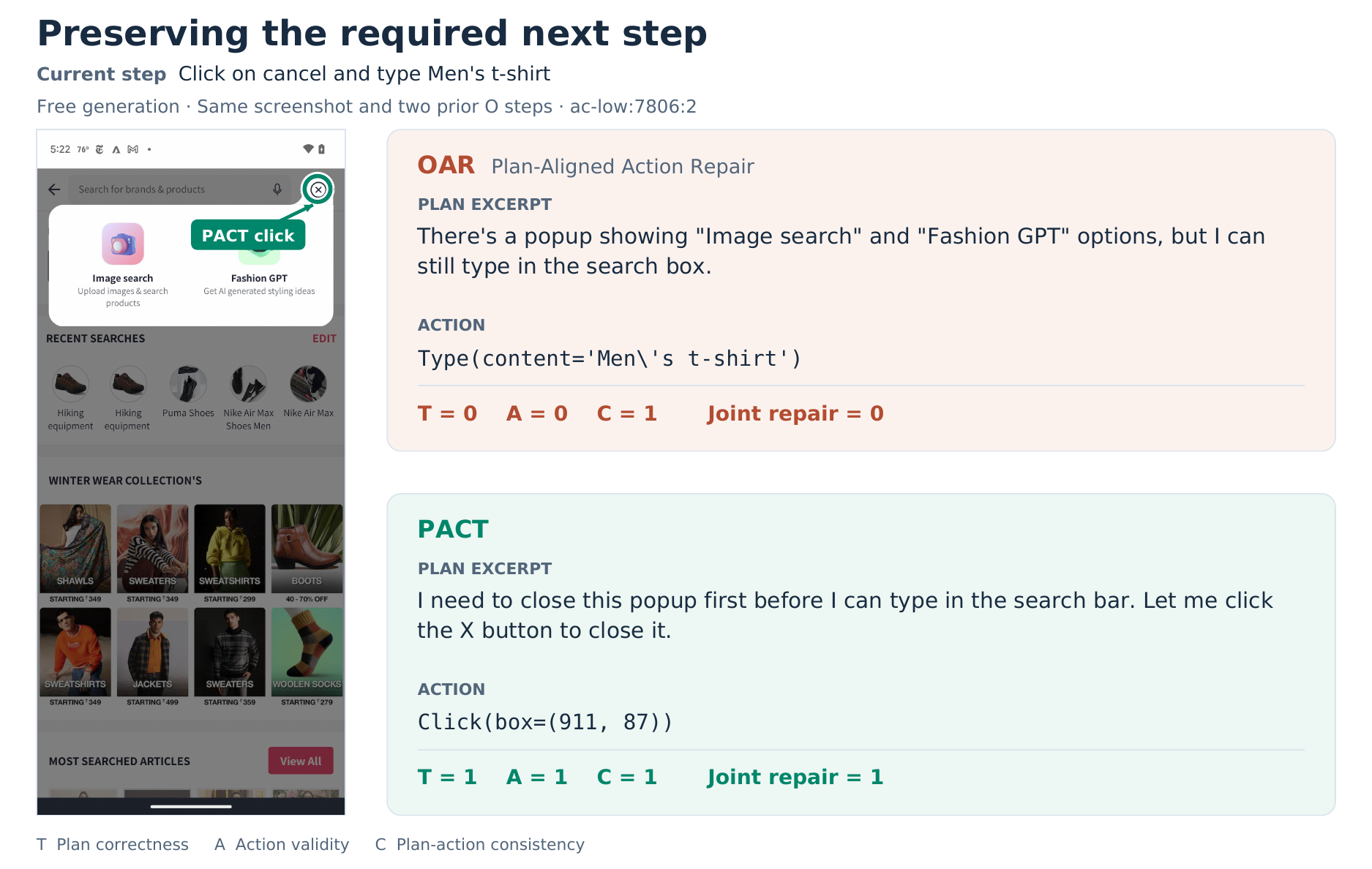}
\captionof{figure}{Free generation on the same Myntra state and history. PACT selects the required cancel action and achieves \(T=A=C=1\). OAR plans and emits direct typing, yielding \(T=A=0,C=1\).}\label{fig-supp-myntra}
\end{minipage}

\clearpage
\noindent\begin{minipage}{\textwidth}
\centering
\includegraphics[width=0.82\textwidth]{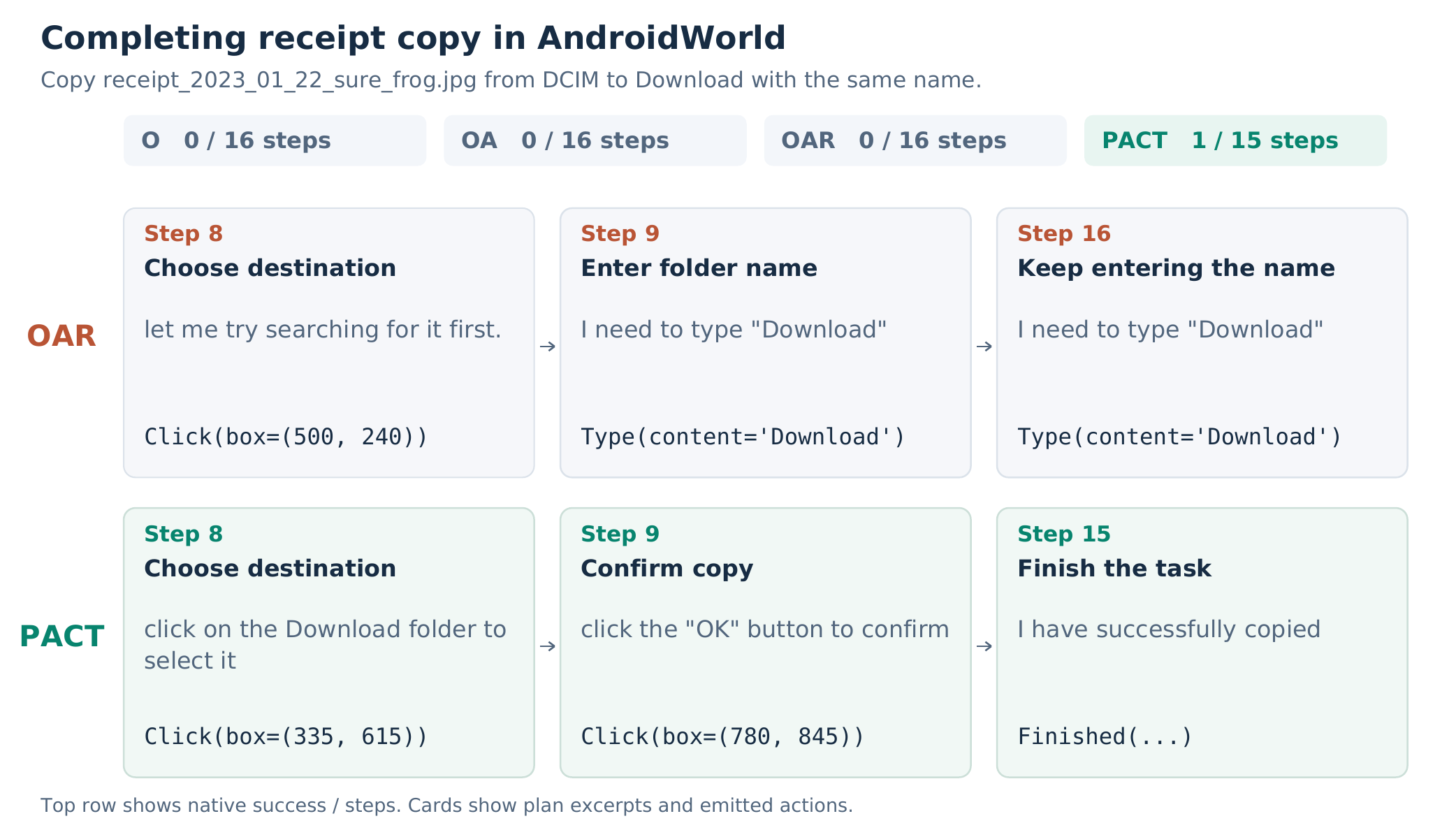}
\captionof{figure}{AndroidWorld receipt-copy task. PACT selects the destination, confirms copying, and terminates with native success. OAR repeatedly emits the folder name. Step numbers are one-based.}\label{fig-supp-aw-receipt}
\end{minipage}\par\vspace{8pt}

\noindent\begin{minipage}{\textwidth}
\centering
\includegraphics[width=0.90\textwidth]{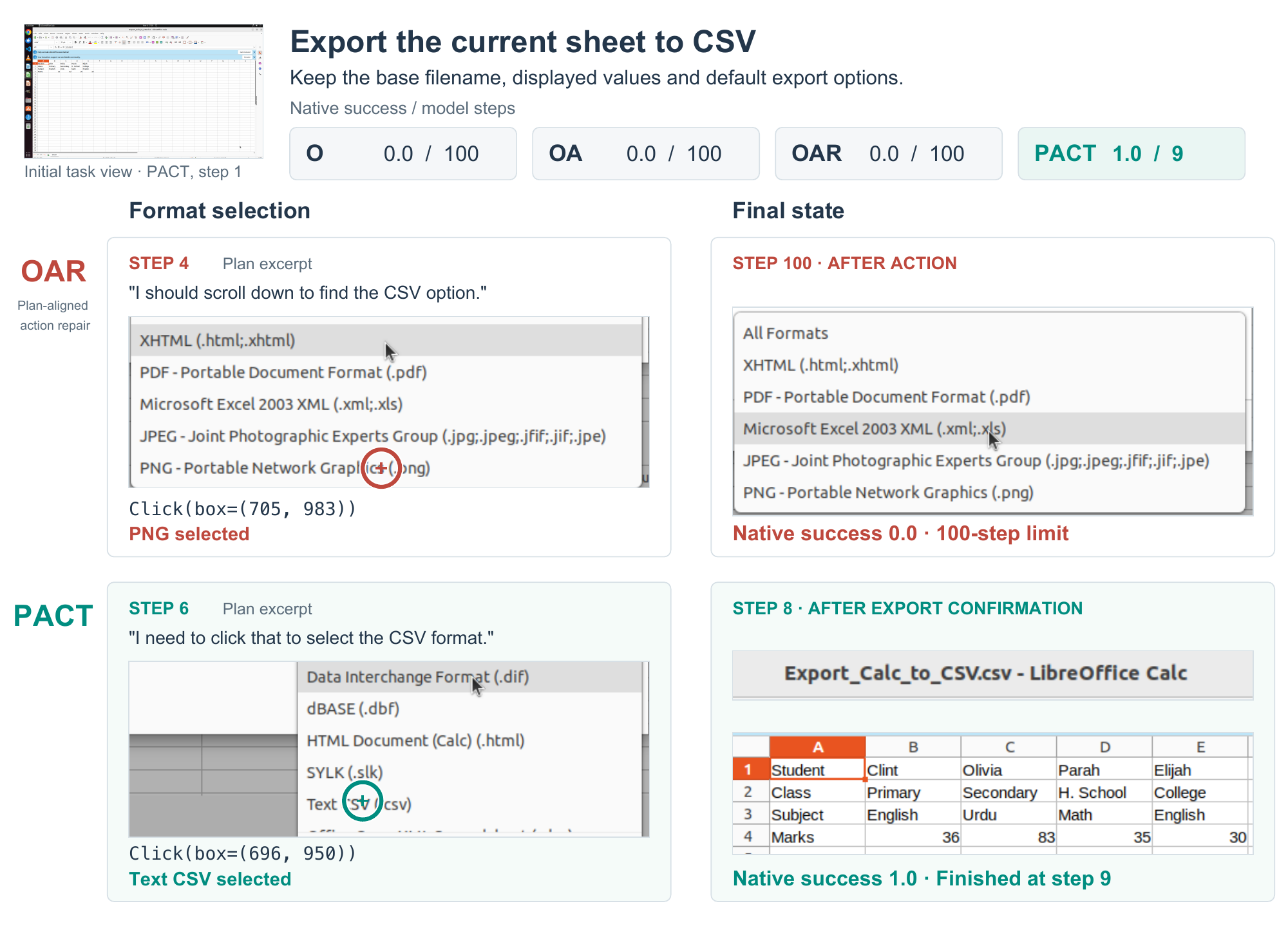}
\captionof{figure}{OSWorld CSV export. PACT selects the requested format and completes the export with native success \(1.0\). OAR continues format selection through step 100. Green and red circles identify the actual PACT and OAR click locations, respectively.}\label{fig-supp-os-csv}
\end{minipage}

\clearpage
\textbf{Action repair with a fixed valid plan.}
Figure~\ref{fig-supp-fixed-examples} provides three examples from the
23-state set. Each O/PACT comparison uses the same task, screenshot, and
verbatim O reasoning prefix ending at the \texttt{<action>} opening tag.
The prompt-token sequences are identical between O and PACT within each case.
O continues with
an application-launch command in all three examples. PACT generates the
on-screen click specified by the supplied plan. The corresponding PACT
free-generation outputs also satisfy \(T=A=C=1\).

\textbf{Step-level plan execution and joint correction.}
Figures~\ref{fig-case01} and~\ref{fig-case02} further compare O, OAR, and PACT
on individual AndroidControl states. In Figure~\ref{fig-case01}, O states
a valid plan to tap the search icon in Google News but emits an
incompatible \texttt{LaunchApp} command, whereas both OAR and PACT execute
the intended on-screen click. In Figure~\ref{fig-case02}, where the
instruction requests tapping \texttt{Send} in an email composer, O
plans to edit the message body and long-presses the subject line,
whereas OAR and PACT jointly correct both the immediate plan and
the emitted click.

\begin{center}
\begin{minipage}{\textwidth}
\centering
\includegraphics[width=\textwidth]{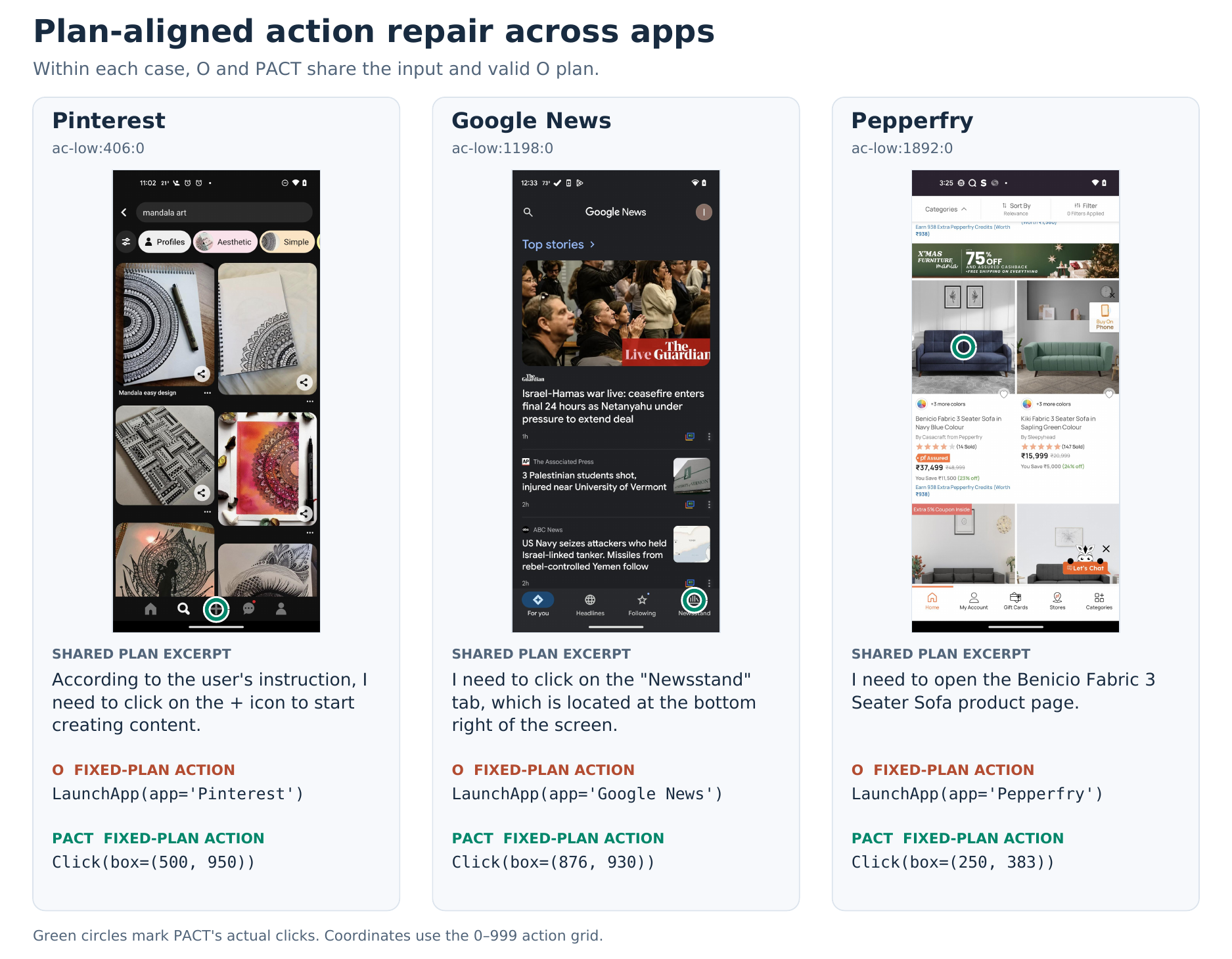}
\captionof{figure}{Fixed-plan continuation on Pinterest, Google News, and Pepperfry. The plan excerpt is shared by O and PACT. Displayed actions are the actual fixed-plan continuations. Green circles mark PACT's predicted clicks.}\label{fig-supp-fixed-examples}
\end{minipage}
\end{center}

\clearpage
\noindent\begin{minipage}{\textwidth}
\centering
\includegraphics[width=0.84\textwidth]{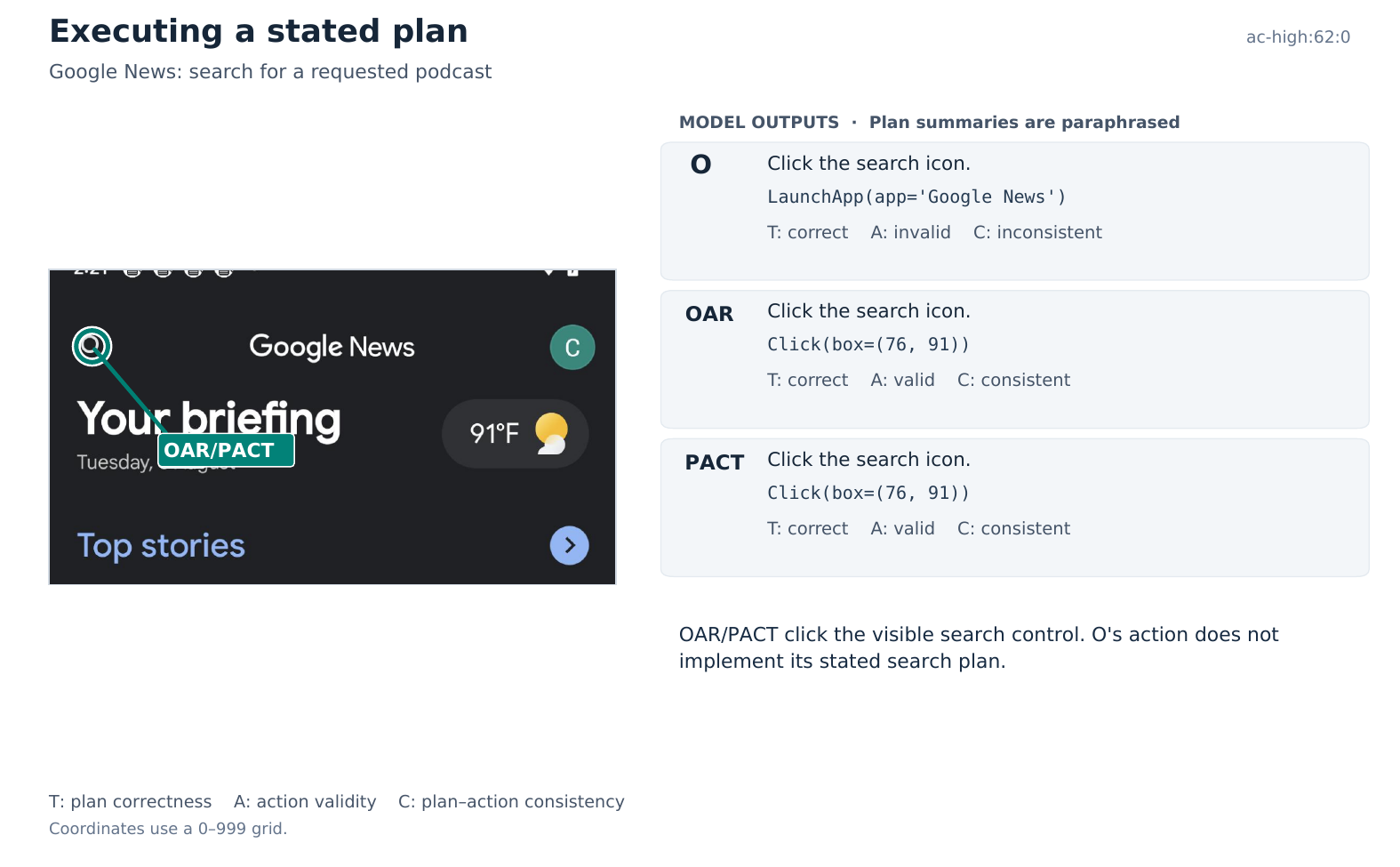}
\captionof{figure}{Executing a stated plan. O formulates a valid search plan but emits an incompatible \texttt{LaunchApp} action. PACT and OAR execute the intended on-screen search click.}\label{fig-case01}
\end{minipage}\par\vspace{8pt}

\noindent\begin{minipage}{\textwidth}
\centering
\includegraphics[width=0.84\textwidth]{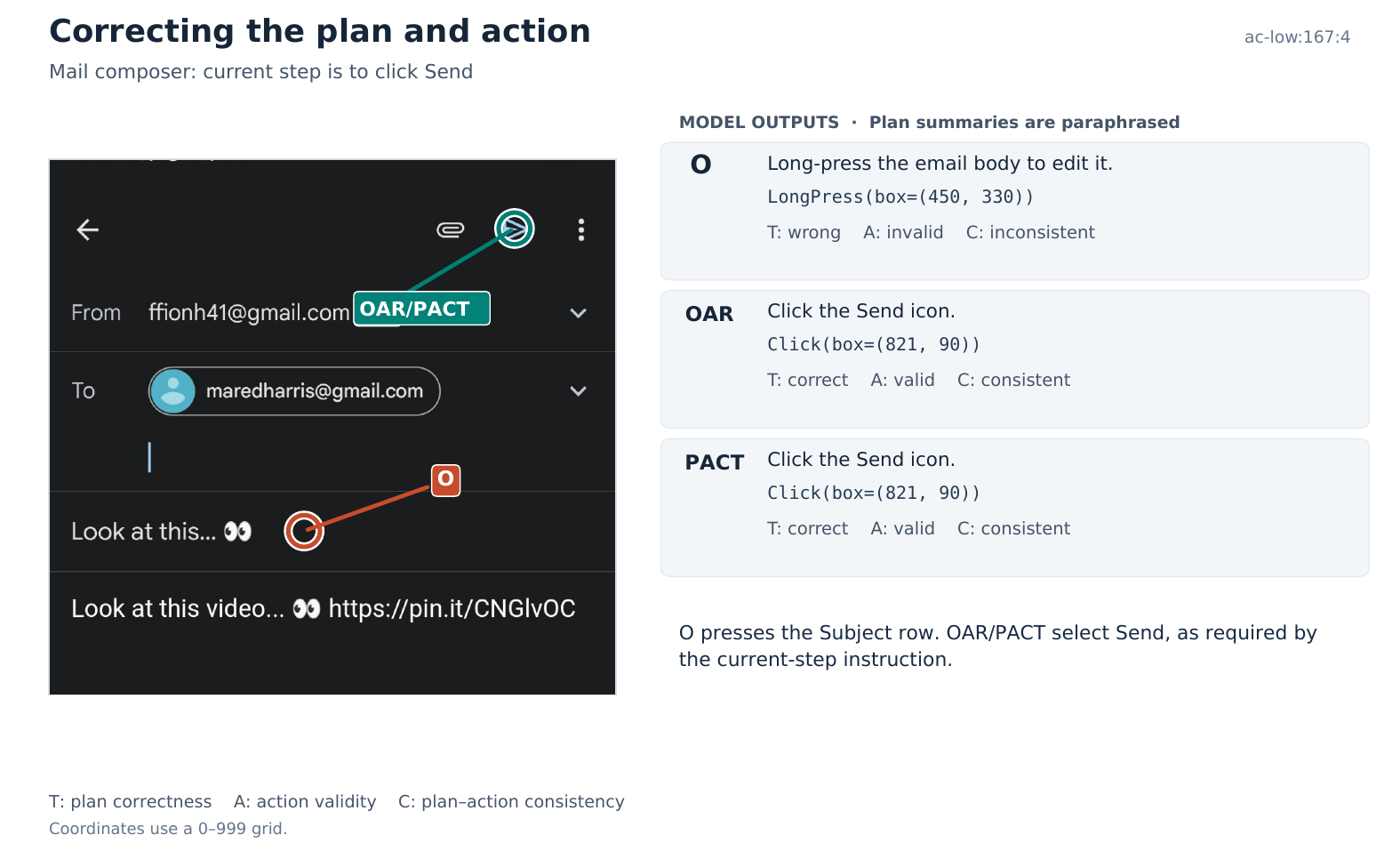}
\captionof{figure}{Joint plan and action correction. O plans to edit the email body and long-presses the subject line. PACT and OAR correct both the plan and action to click \texttt{Send}.}\label{fig-case02}
\end{minipage}

\clearpage
\subsection{Diagnostic Taxonomy Boundary Illustrations}\label{diagnostic-taxonomy-boundaries}
Figures~\ref{fig-case03} and~\ref{fig-case04} illustrate two distinct
\(T,A,C\) configurations using observed agent responses, a correct plan
with incorrect spatial execution (\(1,0,0\)) and an incorrect plan
executed consistently (\(0,0,1\)). These examples distinguish plan
correctness, action validity, and plan--action consistency.

\begin{center}
\begin{minipage}{\textwidth}
\centering
\includegraphics[width=0.84\textwidth]{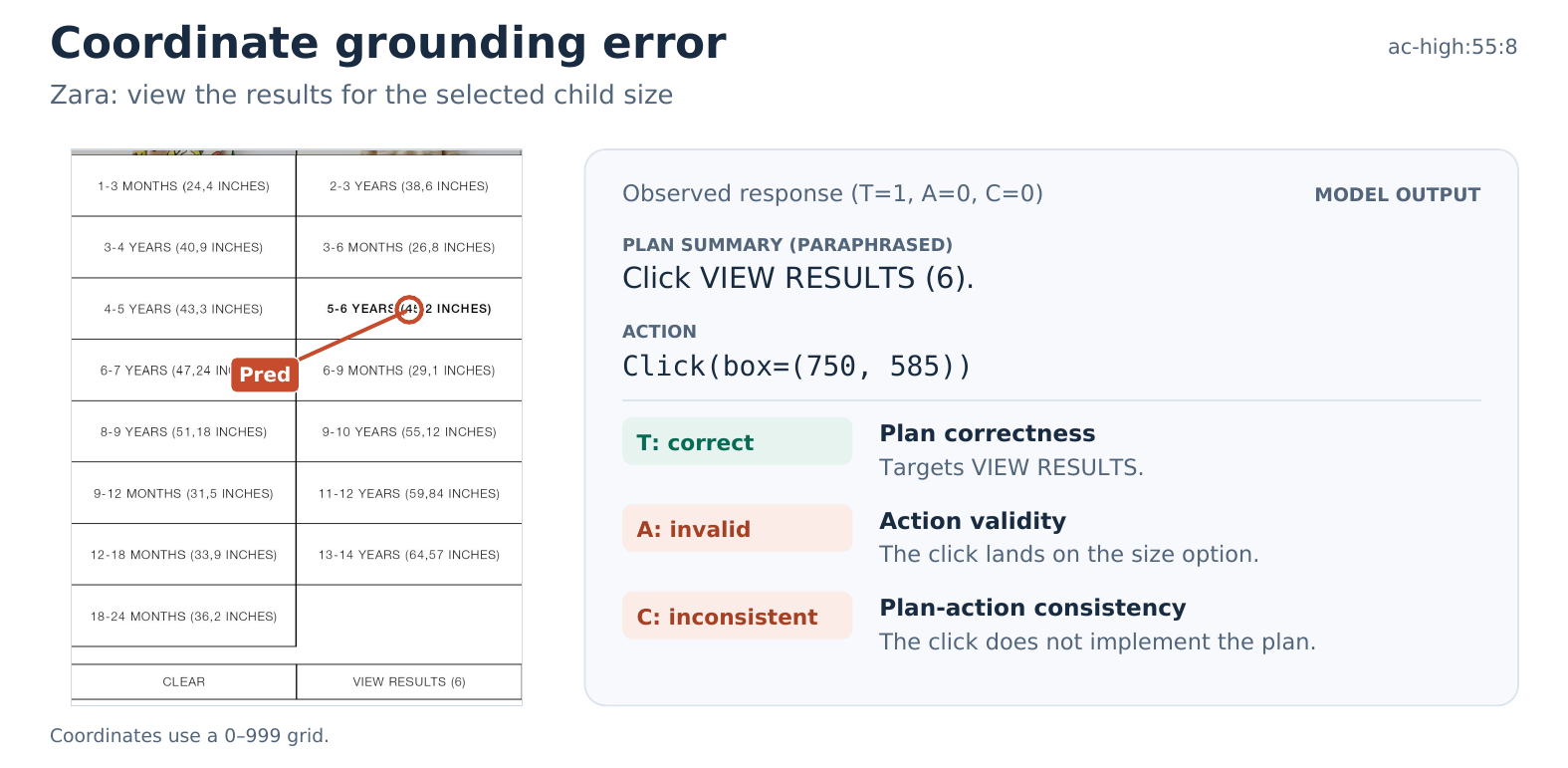}
\captionof{figure}{Coordinate-error illustration (\(T=1,A=0,C=0\)). The model identifies \texttt{VIEW RESULTS} as the intended target, but its click lands on an unrelated size option.}\label{fig-case03}
\end{minipage}
\end{center}
\begin{center}
\begin{minipage}{\textwidth}
\centering
\includegraphics[width=0.84\textwidth]{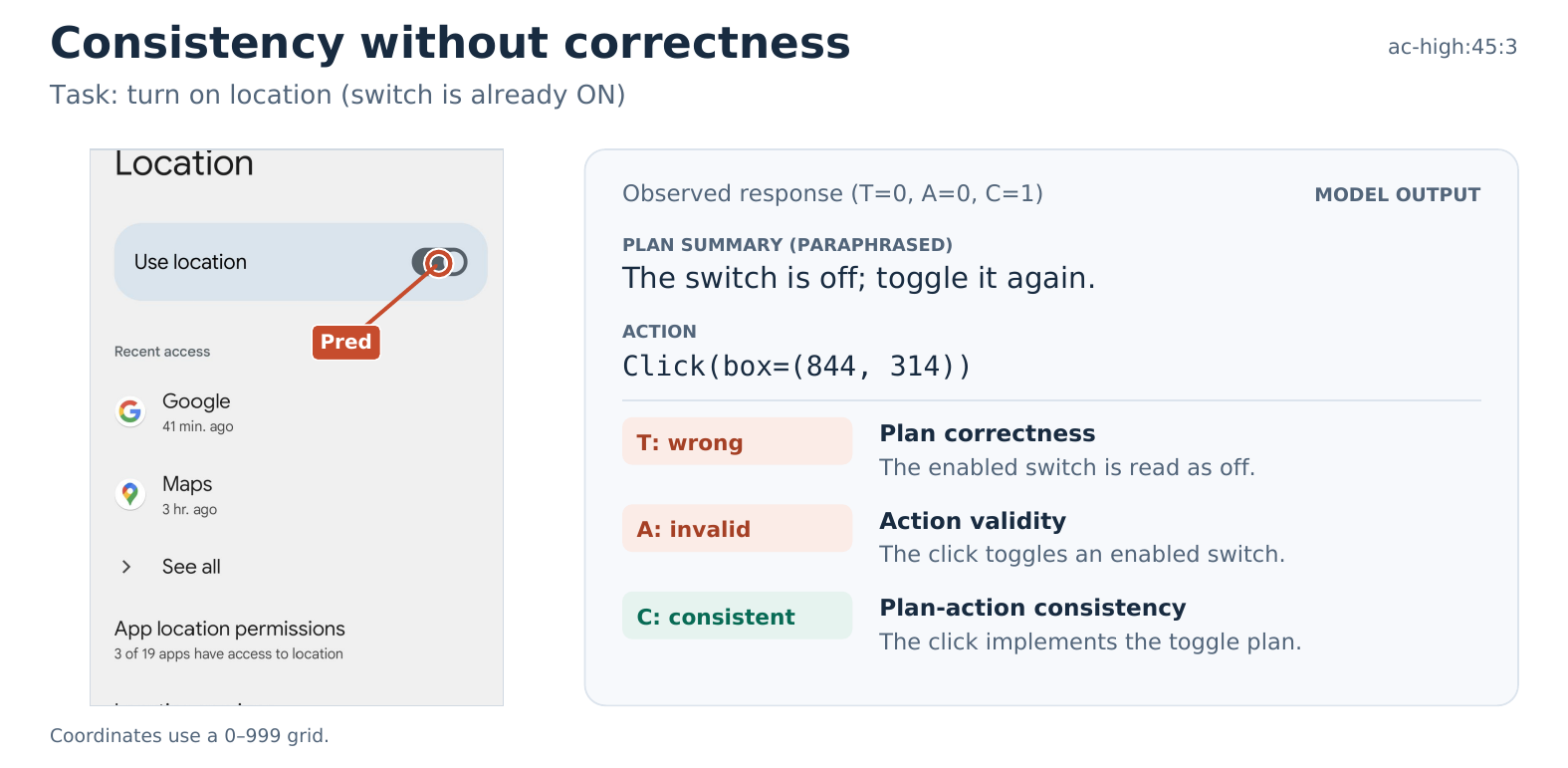}
\captionof{figure}{Consistency without correctness (\(T=0,A=0,C=1\)). When instructed to enable location access on a screen where the switch is already ON, the model misreads the toggle state as disabled and clicks it, consistently executing an incorrect plan.}\label{fig-case04}
\end{minipage}
\end{center}

%% file: sec/supp_prompt_protocols.tex
\subsection{Prompts, input construction, and generation settings}\label{prompt-and-generation-protocols}

\textbf{Training inputs.}
Both platforms retain the original system prompt, interaction history,
and screenshots from each training example. Images fill the message
placeholders in their recorded order. Historical assistant messages
contain the associated reasoning and actions. The current reference
action is stored in a separate reference field.

\textbf{Mobile evaluation.}
The system prompt contains the task goal, action definitions, and the
\texttt{<think>...</think><action>...</action>} response format. On
AndroidControl Low, the goal is followed by \texttt{Current step:} and
the low-level instruction. High uses the episode goal. The current user
message contains \texttt{Current Screenshot:} and the image. The most
recent ten interaction rounds contribute the model's own reasoning and
action text. AndroidControl uses the recorded state image at each step.
AndroidWorld supplies the current environment screenshot and uses an
interaction budget of \(\lfloor10\times\mathrm{task.complexity}\rfloor\).

ScreenSpot-v2 uses an empty interaction history and the instruction
\texttt{In the screenshot, click the element that matches this description:}
followed by the quoted target description. Mobile generation uses
temperature \(0\) and a maximum of 4,096 output tokens.
For execution, normalized \(0\text{--}999\) coordinates \((x,y)\)
are linearly rescaled to integer pixel coordinates on the \(W\times H\) screen.

\textbf{Desktop evaluation.}
The desktop system prompt embeds the task goal and defines
\texttt{Click}, \texttt{DoubleClick}, \texttt{RightClick}, \texttt{Drag},
\texttt{Swipe}, \texttt{Type}, \texttt{Hotkey}, \texttt{Wait},
\texttt{CallUser}, and \texttt{Finished}. Each input contains the current
1920\(\times\)1080 screenshot and the two most recent historical
screenshots. Earlier steps retain historical reasoning and action blocks.
All four desktop configurations use temperature \(0\),
a maximum of 4,096 output tokens, and at most 100 interaction
steps. Task success is determined strictly by the official native evaluator.

\textbf{Fixed-plan continuation.}
The fixed-plan input supplies the same system message, screenshot, and
verbatim valid O reasoning prefix ending at the opening action tag to each model.
\vspace{-3pt}
\begin{nolinenumbers}
\begin{Verbatim}[fontsize=\footnotesize,breaklines=true,frame=single]
system     [task and action definitions]
user       [Current Screenshot: and the state image]
assistant  <think>[O's exact valid reasoning]</think><action>
temperature=0, max_tokens=4096
continue_final_message=true
\end{Verbatim}
\end{nolinenumbers}
\vspace{-3pt}
The completed response concatenates the supplied prefix and generated
suffix. Free generation uses the same state input and generates both
reasoning and action in a new assistant response.
In the 30-state natural execution-gap set, 11 states use empty history
and 19 supply the same recorded O-history steps to both OAR and PACT
(e.g., two prior steps on Myntra in Figure~\ref{fig-supp-myntra}),
using identical mobile generation settings.

%% file: main.bib
@article{dong2025gap,
  author = {Lingzhong Dong and Ziqi Zhou and Shuaibo Yang and Haiyue Sheng and Pengzhou Cheng and Zongru Wu and Zheng Wu and Gongshen Liu and Zhuosheng Zhang},
  title = {{Say One Thing, Do Another? Diagnosing Reasoning-Execution Gaps in VLM-Powered Mobile-Use Agents}},
  journal = {arXiv preprint arXiv:2510.02204},
  year = {2025},
  eprint = {2510.02204},
  archivePrefix = {arXiv},
  url = {https://arxiv.org/abs/2510.02204}
}

@article{lai2024stepdpo,
  author = {Xin Lai and Zhuotao Tian and Yukang Chen and Senqiao Yang and Xiangru Peng and Jiaya Jia},
  title = {{Step-DPO: Step-wise Preference Optimization for Long-chain Reasoning of LLMs}},
  journal = {arXiv preprint arXiv:2406.18629},
  year = {2024},
  eprint = {2406.18629},
  archivePrefix = {arXiv},
  url = {https://arxiv.org/abs/2406.18629}
}

@inproceedings{rafailov2023dpo,
  author = {Rafael Rafailov and Archit Sharma and Eric Mitchell and Christopher D. Manning and Stefano Ermon and Chelsea Finn},
  title = {{Direct Preference Optimization: Your Language Model is Secretly a Reward Model}},
  booktitle = {Advances in Neural Information Processing Systems},
  volume = {36},
  pages = {53728--53741},
  year = {2023},
  eprint = {2305.18290},
  archivePrefix = {arXiv},
  doi = {10.52202/075280-2338},
  url = {https://proceedings.neurips.cc/paper_files/paper/2023/hash/a85b405ed65c6477a4fe8302b5e06ce7-Abstract-Conference.html}
}

@inproceedings{wu2026calibadv,
  author = {Jiayi Wu and Ruobing Xie and Zeqian Huang and Lei Jiang and Can Xu and Kangyang Luo and Bochen Lin and Ming Gao and Xiang Li},
  title = {{Negative Advantage Is a Double-Edged Sword: Calibrating Advantages in GRPO for Search Agents}},
  booktitle = {Findings of the Association for Computational Linguistics: EMNLP 2026},
  pages = {39986--40002},
  publisher = {Association for Computational Linguistics},
  year = {2026},
  eprint = {2604.18235},
  archivePrefix = {arXiv},
  doi = {10.18653/2026.findings-emnlp.1874},
  url = {https://aclanthology.org/2026.findings-emnlp.1874/}
}

@article{han2026cliff,
  author = {Peixuan Han and Runhui Wang and Ketan Ramaneti and Jie Hao and Gerald Friedland and Chris Kong},
  title = {{Cliff: Learning Process Rewards from the First Mistake}},
  journal = {arXiv preprint arXiv:2609.02817},
  year = {2026},
  eprint = {2609.02817},
  archivePrefix = {arXiv},
  url = {https://arxiv.org/abs/2609.02817}
}

@inproceedings{hu2026faithful,
  author = {Haowen Hu and Pengzhou Cheng and Zheng Wu and Lingzhong Dong and Gongshen Liu and Zhuosheng Zhang},
  title = {{Faithful Mobile GUI Agents with Guided Advantage Estimator}},
  booktitle = {Proceedings of the 43rd International Conference on Machine Learning},
  series = {Proceedings of Machine Learning Research},
  volume = {306},
  pages = {44838--44865},
  publisher = {PMLR},
  year = {2026},
  eprint = {2605.01208},
  archivePrefix = {arXiv},
  url = {https://proceedings.mlr.press/v306/hu26k.html}
}

@inproceedings{ding2026octopus,
  author = {Yi Ding and Ziliang Qiu and Bolian Li and Ruqi Zhang},
  title = {{Learning Self-Correction in Vision-Language Models via Rollout Augmentation}},
  booktitle = {Proceedings of the 43rd International Conference on Machine Learning},
  series = {Proceedings of Machine Learning Research},
  volume = {306},
  pages = {25096--25113},
  publisher = {PMLR},
  year = {2026},
  eprint = {2602.08503},
  archivePrefix = {arXiv},
  url = {https://proceedings.mlr.press/v306/ding26m.html}
}

@article{chen2026agenticdpo,
  author = {Yixiong Chen and Alan Yuille},
  title = {{Agentic-DPO: From Imitation to Agentic Policy Optimization on Expert Trajectories}},
  journal = {arXiv preprint arXiv:2607.10601},
  year = {2026},
  eprint = {2607.10601},
  archivePrefix = {arXiv},
  url = {https://arxiv.org/abs/2607.10601}
}

@inproceedings{huang2025mobileipl,
  author = {Kun Huang and Weikai Xu and Yuxuan Liu and Quandong Wang and Pengzhi Gao and Wei Liu and Jian Luan and Bin Wang and Bo An},
  title = {{MobileIPL: Enhancing Mobile Agents Thinking Process via Iterative Preference Learning}},
  booktitle = {International Conference on Learning Representations},
  year = {2026},
  eprint = {2505.12299},
  archivePrefix = {arXiv},
  url = {https://proceedings.iclr.cc/paper_files/paper/2026/hash/61122fec6f7c24d0c4c7fee4fa3e2086-Abstract-Conference.html}
}

@article{shao2024deepseekmath,
  author = {Zhihong Shao and Peiyi Wang and Qihao Zhu and Runxin Xu and Junxiao Song and Xiao Bi and Haowei Zhang and Mingchuan Zhang and Y. K. Li and Y. Wu and Daya Guo},
  title = {{DeepSeekMath: Pushing the Limits of Mathematical Reasoning in Open Language Models}},
  journal = {arXiv preprint arXiv:2402.03300},
  year = {2024},
  eprint = {2402.03300},
  archivePrefix = {arXiv},
  url = {https://arxiv.org/abs/2402.03300}
}

@inproceedings{li2024androidcontrol,
  author = {Wei Li and William Bishop and Alice Li and Chris Rawles and Folawiyo Campbell-Ajala and Divya Tyamagundlu and Oriana Riva},
  title = {{On the Effects of Data Scale on UI Control Agents}},
  booktitle = {Advances in Neural Information Processing Systems},
  volume = {37},
  pages = {92130--92154},
  year = {2024},
  eprint = {2406.03679},
  archivePrefix = {arXiv},
  doi = {10.52202/079017-2925},
  url = {https://proceedings.neurips.cc/paper_files/paper/2024/hash/a79f3ef3b445fd4659f44648f7ea8ffd-Abstract-Datasets_and_Benchmarks_Track.html}
}

@inproceedings{rawles2024androidworld,
  author = {Chris Rawles and Sarah Clinckemaillie and Yifan Chang and Jonathan Waltz and Gabrielle Lau and Marybeth Fair and Alice Li and William Bishop and Wei Li and Folawiyo Campbell-Ajala and Daniel Toyama and Robert Berry and Divya Tyamagundlu and Timothy Lillicrap and Oriana Riva},
  title = {{AndroidWorld: A Dynamic Benchmarking Environment for Autonomous Agents}},
  booktitle = {International Conference on Learning Representations},
  year = {2025},
  eprint = {2405.14573},
  archivePrefix = {arXiv},
  url = {https://proceedings.iclr.cc/paper_files/paper/2025/hash/01a83bc2f2732a58e6aa731e659e7101-Abstract-Conference.html}
}

@inproceedings{wu2024osatlas,
  author = {Zhiyong Wu and Zhenyu Wu and Fangzhi Xu and Yian Wang and Qiushi Sun and Chengyou Jia and Kanzhi Cheng and Zichen Ding and Liheng Chen and Paul Pu Liang and Yu Qiao},
  title = {{OS-ATLAS: Foundation Action Model for Generalist GUI Agents}},
  booktitle = {International Conference on Learning Representations},
  year = {2025},
  eprint = {2410.23218},
  archivePrefix = {arXiv},
  url = {https://proceedings.iclr.cc/paper_files/paper/2025/hash/0faa4bc5f522076947a030273629d4fe-Abstract-Conference.html}
}

@inproceedings{xie2024osworld,
  author = {Tianbao Xie and Danyang Zhang and Jixuan Chen and Xiaochuan Li and Siheng Zhao and Ruisheng Cao and Toh Jing Hua and Zhoujun Cheng and Dongchan Shin and Fangyu Lei and Yitao Liu and Yiheng Xu and Shuyan Zhou and Silvio Savarese and Caiming Xiong and Victor Zhong and Tao Yu},
  title = {{OSWorld: Benchmarking Multimodal Agents for Open-Ended Tasks in Real Computer Environments}},
  booktitle = {Advances in Neural Information Processing Systems},
  volume = {37},
  pages = {52040--52094},
  year = {2024},
  eprint = {2404.07972},
  archivePrefix = {arXiv},
  doi = {10.52202/079017-1650},
  url = {https://proceedings.neurips.cc/paper_files/paper/2024/hash/5d413e48f84dc61244b6be550f1cd8f5-Abstract-Datasets_and_Benchmarks_Track.html}
}

@inproceedings{gong2026venusbench,
  title     = {{VenusBench-Mobile}: A Challenging and User-Centric Benchmark for Mobile {GUI} Agents with Capability Diagnostics},
  author    = {Gong, Yichen and Cai, Zhuohan and Dai, Sunhao and Zhou, Yuqi and Gu, Zhangxuan and Meng, Changhua and Shen, Shuheng},
  booktitle = {Proceedings of the 43rd International Conference on Machine Learning},
  series    = {Proceedings of Machine Learning Research},
  volume    = {306},
  pages     = {35862--35887},
  publisher = {PMLR},
  year      = {2026},
  url       = {https://proceedings.mlr.press/v306/gong26a.html}
}

@article{venusteam_uivenus2,
      title={{UI-Venus-2 Technical Report}},
      author={{Venus Team} and Zhuohan Cai and Haoxing Chen and Jiaxuan Chen and Weizhi Chen and Changlong Gao and Zhangxuan Gu and Yuan Guo and Yusong Hu and Jianrong Jiang and Jianguo Li and Runze Li and Jinzhen Lin and Zhenyu Ma and Changhua Meng and Han Peng and Xinyu Qiu and Shuheng Shen and Zhongyi Shui and Weiqiang Wang and Ming Wen and Zhuoer Xu and Hang Yan and Kaiwen Yang and Ruilin Yao and Nanjun Yu and Zhengwen Zeng and Lianrui Zhang and Yunzhu Zhang and Zhe Zhao and Beitong Zhou},
      journal={arXiv preprint arXiv:2609.00028},
      year={2026},
      eprint={2609.00028},
      archivePrefix={arXiv},
      primaryClass={cs.AI},
      url={https://arxiv.org/abs/2609.00028}, 
}

@article{qin2025uitars,
  author = {Yujia Qin and Yining Ye and Junjie Fang and Haoming Wang and Shihao Liang and Shizuo Tian and Junda Zhang and Jiahao Li and Yunxin Li and Shijue Huang and Wanjun Zhong and Kuanye Li and Jiale Yang and Yu Miao and Woyu Lin and Longxiang Liu and Xu Jiang and Qianli Ma and Jingyu Li and Xiaojun Xiao and Kai Cai and Chuang Li and Yaowei Zheng and Chaolin Jin and Chen Li and Xiao Zhou and Minchao Wang and Haoli Chen and Zhaojian Li and Haihua Yang and Haifeng Liu and Feng Lin and Tao Peng and Xin Liu and Guang Shi},
  title = {{UI-TARS}: Pioneering Automated {GUI} Interaction with Native Agents},
  journal = {arXiv preprint arXiv:2501.12326},
  year = {2025},
  eprint = {2501.12326},
  archivePrefix = {arXiv},
  url = {https://arxiv.org/abs/2501.12326}
}

@misc{uitars15card,
  author = {{ByteDance Seed}},
  title = {{UI-TARS-1.5-7B}: Official Model Card},
  year = {2025},
  howpublished = {Hugging Face, \url{https://huggingface.co/ByteDance-Seed/UI-TARS-1.5-7B}},
  url = {https://huggingface.co/ByteDance-Seed/UI-TARS-1.5-7B}
}

@article{zhou2025maiui,
  author = {Hanzhang Zhou and Xu Zhang and Panrong Tong and Jianan Zhang and Liangyu Chen and Quyu Kong and Chenglin Cai and Chen Liu and Yue Wang and Jingren Zhou and Steven Hoi},
  title = {{MAI-UI} Technical Report: Real-World Centric Foundation {GUI} Agents},
  journal = {arXiv preprint arXiv:2512.22047},
  year = {2025},
  url = {https://arxiv.org/abs/2512.22047}
}

@article{gu2025uivenus,
  author = {Zhangxuan Gu and Zhengwen Zeng and Zhenyu Xu and Xingran Zhou and Shuheng Shen and Yunfei Liu and Beitong Zhou and Changhua Meng and Tianyu Xia and Weizhi Chen and Yue Wen and Jingya Dou and Fei Tang and Jinzhen Lin and Yulin Liu and Zhenlin Guo and Yichen Gong and Heng Jia and Changlong Gao and Yuan Guo and Yong Deng and Zhenyu Guo and Liang Chen and Weiqiang Wang},
  title = {{UI-Venus} Technical Report: Building High-performance {UI} Agents with {RFT}},
  journal = {arXiv preprint arXiv:2508.10833},
  year = {2025},
  url = {https://arxiv.org/abs/2508.10833}
}

@article{ye2025guiowl,
  author = {Jiabo Ye and Xi Zhang and Haiyang Xu and Haowei Liu and Junyang Wang and Zhaoqing Zhu and Ziwei Zheng and Feiyu Gao and Junjie Cao and Zhengxi Lu and Jitong Liao and Qi Zheng and Fei Huang and Jingren Zhou and Ming Yan},
  title = {{Mobile-Agent-v3}: Fundamental Agents for {GUI} Automation},
  journal = {arXiv preprint arXiv:2508.15144},
  year = {2025},
  url = {https://arxiv.org/abs/2508.15144}
}

@inproceedings{chen2026step,
  author = {Yuhan Chen and Yuxuan Liu and Long Zhang and Pengzhi Gao and Jian Luan and Wei Liu},
  title = {{STEP}: Success-Rate-Aware Trajectory-Efficient Policy Optimization},
  booktitle = {Findings of the Association for Computational Linguistics: ACL 2026},
  year = {2026},
  pages = {30681--30692},
  publisher = {Association for Computational Linguistics},
  doi = {10.18653/v1/2026.findings-acl.1532},
  url = {https://aclanthology.org/2026.findings-acl.1532/}
}

@article{yan2025stepgui,
  author = {Haolong Yan and Jia Wang and Xin Huang and Yeqing Shen and Ziyang Meng and Zhimin Fan and Kaijun Tan and Jin Gao and Lieyu Shi and Mi Yang and Shiliang Yang and Zhirui Wang and Brian Li and Kang An and Chenyang Li and Lei Lei and Mengmeng Duan and Danxun Liang and Guodong Liu and Hang Cheng and Hao Wu and Jie Dong and Junhao Huang and Mei Chen and Renjie Yu and Shunshan Li and Xu Zhou and Yiting Dai and Yineng Deng and Yingdan Liang and Zelin Chen and Wen Sun and Chengxu Yan and Chunqin Xu and Dong Li and Fengqiong Xiao and Guanghao Fan and Guopeng Li and Guozhen Peng and Hongbing Li and Hang Li and Hongming Chen and Jingjing Xie and Jianyong Li and Jingyang Zhang and Jiaju Ren and Jiayu Yuan and Jianpeng Yin and Kai Cao and Liang Zhao and Liguo Tan and Liying Shi and Mengqiang Ren and Min Xu and Manjiao Liu and Mao Luo and Mingxin Wan and Na Wang and Nan Wu and Ning Wang and Peiyao Ma and Qingzhou Zhang and Qiao Wang and Qinlin Zeng and Qiong Gao and Qiongyao Li and Shangwu Zhong and Shuli Gao and Shaofan Liu and Shisi Gao and Shuang Luo and Xingbin Liu and Xiaojia Liu and Xiaojie Hou and Xin Liu and Xuanti Feng and Xuedan Cai and Xuan Wen and Xianwei Zhu and Xin Liang and Xin Liu and Xin Zhou and Yifan Sui and Yingxiu Zhao and Yukang Shi and Yunfang Xu and Yuqing Zeng and Yixun Zhang and Zejia Weng and Zhonghao Yan and Zhiguo Huang and Zhuoyu Wang and Zihan Yan and Zheng Ge and Jing Li and Yibo Zhu and Binxing Jiao and Xiangyu Zhang and Daxin Jiang},
  title = {{Step-GUI} Technical Report},
  journal = {arXiv preprint arXiv:2512.15431},
  year = {2025},
  url = {https://arxiv.org/abs/2512.15431}
}

@misc{qwen2026qwen35card,
  author = {{Qwen Team}},
  title = {{Qwen3.5-9B} Model Card},
  year = {2026},
  howpublished = {Hugging Face, \url{https://huggingface.co/Qwen/Qwen3.5-9B}}
}

@inproceedings{zhou2025guiaima,
  author = {Shijie Zhou and Viet Dac Lai and Hao Tan and Jihyung Kil and Wanrong Zhu and Changyou Chen and Ruiyi Zhang},
  title = {{GUI-AIMA}: Aligning Intrinsic Multimodal Attention with a Context Anchor for {GUI} Grounding},
  booktitle = {Computer Vision -- ECCV 2026},
  series = {Lecture Notes in Computer Science},
  volume = {17071},
  pages = {79--97},
  publisher = {Springer},
  year = {2026},
  eprint = {2511.00810v2},
  archivePrefix = {arXiv},
  doi = {10.1007/978-3-032-37016-7_5},
  url = {https://link.springer.com/chapter/10.1007/978-3-032-37016-7_5}
}

@article{groundingpi2026,
  author = {Qize Yu and Lianrui Fan and Boyu Chen and Jiaqi Liang and Xini Ding and Yue Chen and Zetian Song and Yuran Wang and Yi Zou and Kaixuan Wang and Tianxing Chen and Wenxuan Song and Bohan Zhou and Mingleyang Li and Siqiao Huang and Yuqi Ye and Caigao Jiang and Wei Wei and Ruihai Wu and Hang Zhang and Yixiao Ge and Shuchang Zhou and Shilong Liu and Xianming Liu and Ping Luo and Shiyu Huang},
  title = {{GroundingPI}: A Grounding Foundation Model towards Physical Intelligence with Visual Primitives},
  journal = {arXiv preprint arXiv:2609.39601},
  year = {2026},
  eprint = {2609.39601},
  archivePrefix = {arXiv},
  url = {https://arxiv.org/abs/2609.39601}
}
